\documentclass[letterpaper]{article} 
\usepackage{aaai2026}  
\usepackage{times}  
\usepackage{helvet}  
\usepackage{courier}  
\usepackage[hyphens]{url}  
\usepackage{graphicx} 
\usepackage{natbib}  
\usepackage{caption} 
\usepackage{algorithm}
\usepackage{algorithmic}

\usepackage{newfloat}
\usepackage{listings}
\DeclareCaptionStyle{ruled}{labelfont=normalfont,labelsep=colon,strut=off} 
\floatstyle{ruled}
\newfloat{listing}{tb}{lst}{}
\floatname{listing}{Listing}
\usepackage{amsmath}
\usepackage{amssymb}
\usepackage{mathtools}
\usepackage{amsthm}
\usepackage{graphicx}
\usepackage{algorithmic}
\usepackage{algorithm}
\usepackage{subcaption}
\usepackage{url}
\usepackage{booktabs}
\usepackage{tabularx}

\usepackage{booktabs}
\usepackage{longtable}

\usepackage{listings}
\usepackage{xcolor}

\lstdefinestyle{python}{
    backgroundcolor=\color{lightgray},   
    commentstyle=\color{green},
    keywordstyle=\color{blue},
    numberstyle=\tiny\color{gray},
    stringstyle=\color{red},
    basicstyle=\ttfamily,
    breaklines=true,                     
    numbers=left,                        
    numbersep=5pt,                       
    showstringspaces=false,
}
\theoremstyle{plain}
\newtheorem{theorem}{Theorem}[section]

\newtheorem{corollary}[theorem]{Corollary}
\theoremstyle{definition}
\newtheorem{definition}[theorem]{Definition}

\theoremstyle{remark}

\title{EvoGrad: Evolutionary-Weighted Gradient and Hessian Learning \\ for Black-Box Optimization}
\author {
    Yedidya Kfir \textsuperscript{\rm 1},
    Elad Sarafian\textsuperscript{\rm 2},
    Yoram Louzoun\textsuperscript{\rm 1},
    Sarit Kraus\textsuperscript{\rm 1}
}
\affiliations {
    \textsuperscript{\rm 1}Bar-Ilan University, 
    \textsuperscript{\rm 2}Nvidia \\
    kfiry@biu.ac.il, esarafian@nvidia.com, louzouy@math.biu.ac.il, sarit@cs.biu.ac.il
}

\usepackage{bibentry}

\begin{document}

\maketitle

\begin{abstract}
\begin{quote}

Black-box algorithms aim to optimize functions without access to their analytical structure or gradient information, making them essential when gradients are unavailable or computationally expensive to obtain. Traditional methods for black-box optimization (BBO) primarily utilize non-parametric models, but these approaches often struggle to scale effectively in large input spaces. Conversely, parametric approaches, which rely on neural estimators and gradient signals via backpropagation, frequently encounter substantial gradient estimation errors, limiting their reliability. Explicit Gradient Learning (EGL), a recent advancement, directly learns gradients using a first-order Taylor approximation and has demonstrated superior performance compared to both parametric and non-parametric methods. However, EGL inherently remains local and myopic, often faltering on highly non-convex optimization landscapes. In this work, we address this limitation by integrating global statistical insights from the evolutionary algorithm CMA-ES into the gradient learning framework, effectively biasing gradient estimates towards regions with higher optimization potential. Moreover, we enhance the gradient learning process by estimating the Hessian matrix, allowing us to correct the second-order residual of the Taylor series approximation. Our proposed algorithm, EvoGrad\textsuperscript{2} (Evolutionary Gradient Learning with second-order approximation), achieves state-of-the-art results on the synthetic COCO test suite, exhibiting significant advantages in high-dimensional optimization problems. We further demonstrate EvoGrad\textsuperscript{2}'s effectiveness on challenging real-world machine learning tasks, including adversarial training and code generation, highlighting its ability to produce more robust, high-quality solutions. Our results underscore EvoGrad\textsuperscript{2}'s potential as a powerful tool for researchers and practitioners facing complex, high-dimensional, and non-linear optimization problems.

\end{quote}
\end{abstract}


\section{Introduction} \label{sec:intro}
Black-box optimization (BBO) is the process of searching for optimal solutions within a system's input domain without access to its internal structure or analytical properties \cite{audet2017introduction}. Unlike gradient-based optimization methods that rely on the calculation of analytical gradients, BBO algorithms query the system solely through input-output pairs, operating agnostically to the underlying function. This feature distinguishes BBO from traditional ML tasks, such as neural network training, where optimization typically involves backpropagation-based gradient computation.

Many real-world systems naturally fit the BBO framework because their behavior is difficult or impossible to model explicitly. In such cases, BBO algorithms have achieved remarkable success across fields, such as ambulance deployment \cite{zhen2014simulation}, robotic motor control \cite{gehring2014towards,prabhu2018survey}, parameter tuning \cite{olof2018comparative,rimon2024mamba}, and signal processing \cite{zoReview}, among others \cite{alarie2021two}. BBO applications also go beyond physical systems; many ML problems are black-box when the true gradient is unavailable. Examples include hyperparameter tuning \cite{bischl2023hyperparameter}, contextual bandit problems \cite{bouneffouf2020survey}, large language model training with human feedback \cite{bai2022training}, and optimizing language models for long context inference \cite{ding2024longrope}, to name a few.


As problem dimensions increase, the cost of evaluating intermediate solutions becomes a critical constraint, especially in real-world settings where interaction with the environment is expensive or in ML, where larger models demand substantial computational power. Therefore, modern BBO algorithms must reduce evaluation steps \cite{hansen2010comparing}. Achieving this requires algorithms capable of more accurately predicting optimization directions, either through better gradient approximation \cite{anil2020scalable,lesage2020second} or momentum-based strategies to handle non-convexity and noise. In this paper, we propose two novel methods: (1) Evolutionary Gradient Learning (EvoGrad), a weighted gradient estimator that biases toward promising solutions, and (2) Higher-Order Gradient Learning (HGrad), which incorporates Hessian corrections to yield more accurate gradient approximations.

We unify the strengths of EvoGrad and HGrad in EvoGrad\textsuperscript{2}, which offers four key advantages:

\begin{itemize}
    \item \textbf{Performance}: EvoGrad\textsuperscript{2} consistently outperforms baseline algorithms across a diverse range of problems. Including synthetic test suites and real-world ML applications. It can handle noisy and non-convex environments.
    \item \textbf{Strategic Global Gradient Estimation}: Although traditional local search methods can become trapped due to exploitation steps, our algorithm strategically incorporates statistical biases toward promising global regions, enhancing the likelihood of identifying directions leading closer to the global minimum.
    \item \textbf{Success Rate}: Our algorithms are consistent throughout the suite, able to solve more problems than other algorithms on the benchmark.
\end{itemize}

\textbf{Related works:}
Black-box optimization (BBO) algorithms have a long history, with various approaches developed over the years. Some of the foundational techniques include grid search, coordinate search \cite{audet2017searchmethods}, simulated annealing \cite{busetti2003simulated}, and direct search methods like Generalized Pattern Search and Mesh Adaptive direct search \cite{audet2017direct}, Gradient-less descent \cite{golovin2019gradientless}, and ZOO \cite{chen2017zoo}. These approaches iteratively evaluate potential solutions and decide whether to continue in the same direction. However, they resample at every step and don't reuse the budget from previous iterations, wasting a lot of budget.

Another prominent family of BBO algorithms is the evolutionary methods \cite{back1996evolutionary}. Including methods such as Covariance Matrix Adaptation (CMA) \cite{hansen2016cma} and Particle Swarm Optimization (PSO) \cite{clerc2010particle}. These simulate the process of natural evolution, where a population of solutions evolves through mutation and selection \cite{audet2017genetic}. 
They perform well in BBO environments due to their effectiveness in tackling non-convex problems.
However, they come with significant drawbacks, particularly the need for extensive fine-tuning of parameters like generation size and mutation rates. CMA, for example, struggles in higher-dimensional environments and requires careful adjustment of hyperparameters and guidance to perform optimally \cite{loshchilov2013bi,tang2021guiding}.
Recently, \cite{braun2024stein} uses CMA as a building block for SVGD \cite{liu2016stein} in intractable environments, highlighting the diverse ways in which CMA and evolution strategies can be repurposed.

Then there are model-based methods \cite{audet2017model}, which attempt to emulate the behavior of the function using a surrogate model. These models provide important analytical information, such as gradients \cite{bertsekas2015convex}, to guide the optimization process and help find a minimum. Within this class, we can further distinguish two sub-classes.
To address the issue of dimensionality, Explicit Gradient Learning (EGL) was proposed by \cite{sarafian2020explicit}. While many model-based methods focus on learning the function’s structure to derive analytical insights (e.g., Indirect Gradient Learning or IGL \cite{lillicrap2015continuous,sarafian2020explicit}), EGL directly learns the gradient information. EGL uses Taylor's theorem to estimate the gradient. The authors also emphasize the importance of utilizing a trust region to handle black-box optimization problems. 


Several works have explored hybrid approaches that combine gradient-based methods with evolutionary algorithms, aiming to leverage both global search and local descent capabilities. For example, \cite{liu2020self,maheswaranathan2019guided,kunpeng2012descent} utilize gradient information to construct probabilistic models or distributions, which are then employed to estimate the search direction. On the other hand, \cite{tang2021guiding} took a different approach. Ignoring the functions' gradient information, it seeks to train a generative model from the function’s distribution and generate better candidate solutions from the model.


\textbf{The paper is organized as follows:} Section \ref{sec:background} covers the algorithm’s theoretical background and mathematical foundations. Sections \ref{sec:OGL} and \ref{sec:improvements} present our two enhanced variants of the gradient learning algorithm: EvoGrad and HGrad. These are followed by section \ref{sec:findings} where we present the full algorithm, EvoGrad\textsuperscript{2}\footnote{https://github.com/yedidyakfir/egl}. Section \ref{sec:experiment} provides experimental results on the synthetic COCO test suite, and Section \ref{sec:applications} highlights 2 real-world high-dimensional applications and potential uses. Finally, section \ref{sec:conclusion} concludes and suggests future research directions. The supplementary material shows our code, experiments, and environment setup.

\section{Background} \label{sec:background}
\textbf{BBO:}
The goal of black-box optimization (BBO) is to minimize a target function \(f(x)\) through a series of evaluations \cite{audet2017introduction}, over a predefined domain $\Omega$:
\begin{equation}
    \text{find:}\ \  x^\ast=\arg\min_{x \in \Omega}f(x)
\end{equation}
\textbf{Explicit Gradient Learning:} 
The Explicit Gradient Learning method, as proposed by \cite{sarafian2020explicit}, leverages the first-order Taylor’s expansion: $f(y) = f(x) + \nabla f(x)^\top(y-x) + R_1(x, y)$. Here, $R_1(x, y)=O(\|y-x\|^2)$ is a higher-order residual. By minimizing the residual term with a surrogate neural network model, EGL learns the \textit{mean-gradient}: a smooth approximation of the function's gradient
\begin{equation}
\begin{aligned}
\label{eq:EGL_surrogate}
 g^{EGL}_{\varepsilon}(x) = \arg\min_{g_{\theta}:\mathbb{R}^n \rightarrow \mathbb{R}^n}\int_{\tau \in B_\varepsilon(0)} (\mathcal{R}_{g_\theta, x}^{EGL}(\tau))^2 d\tau\\
\mathcal{R}_{g_\theta, x}^{EGL}(\tau) = f(x) - f(x + \tau) + g_{\theta}(x)^\top \tau
\end{aligned}
\end{equation}

Here, \(B_\varepsilon(0)\) is a ball around $0$ which defines the acquisition region of new statistics, where usually new samples are uniformly sampled from this region. As \(\varepsilon\to 0\), the mean-gradient converges to $\nabla f$. EGL thus uses \(\varepsilon\) to control the accuracy of the mean-gradient. This property enables EGL to explore the entire landscape. Specifically, when \(\varepsilon\) is sufficiently large, EGL can locate lower regions in the function. Conversely, when \(\varepsilon \to 0\), it converges to a local minimum. Leveraging this property, the paper shows that it is possible to determine both the algorithm's convergence rate and the estimated gradient's error rate. See corollary \ref{eq:convergence_base_collary}.

\textbf{CMA-ES:} Covariance Matrix Adaptation Evolution Strategy (CMA-ES) builds and updates a multivariate Gaussian distribution to guide optimization. The key element is the covariance matrix, which captures the shape and orientation of the search distribution based on selected samples.

For each generation $g$, the algorithm samples $x_1,..., x_{\lambda} \sim \mathcal{N}(m, \sigma^2C)$, where $m,\sigma$ and $C$ are the evolving mean, standard deviation and covariance matrix parameters.
The sampled points are then sorted according to their \textit{fitness} function values, i.e. $[x_1,...x_\lambda]$ s.t. $(f(x_i) \leq f(x_j): \forall i < j$, and a set of recombination weights $w_1, \dots, w_n$ is assigned for each individual. These weights are positive, decreasing with rank ($w_1 \geq w_2 \geq \dots \geq w_n > 0$), and normalized such that $\sum_{i=1}^\mu w_i = 1$.
These samples are then used to update an improve the evolved parameters $m,\sigma$ and $C$ for the next generation $g+1$. For example, the covariance matrix is updated by first calculating and estimated statistics of the $g$ generation covariance:
\[
C' = \sum_{i=1}^\mu w_i \cdot \left( \frac{x_{i} - m}{\sigma} \right) \left( \frac{x_{i} - m}{\sigma} \right)^\top\
\]
and then applying a weighted update to construct the new generation covariance matrix
\[C^g = (1 - c_\mu) C^{g-1} + c_\mu C'\].
Here $C^{g-1}$ and $C^g$ are the previous and new generation matrices and $c_\mu$ is a hyperparameter. Likewise $m$ and $\sigma$ are updated to improved the population statistics and converge to the optimal solution (usually defined as $m$ or $x_1$ of the last generation).

\textbf{Trust Region:} A powerful tool for BBO algorithms is a trust region (TR), a mechanism that restricts the search to a localized area around the current estimate. 
By constraining the optimization steps and standardizing the input and output statistics. \cite{sarafian2020explicit} showed the usefulness of this tool when applying gradient learning with neural-networks. In our work, we applied TR also to CMA-ES to create strong baseline algorithms, see Appendix: Algorithm \ref{alg:CMA_TR_ALG}. This modification, though conceptually simple, leads to significant performance gains, especially in high-dimensional settings.
\section{Gradient Learning with Evolutionary Weights}
\label{sec:OGL}
Local search algorithms are based on the notion that the gradient descent path is the optimal search path. However, this assumption often proves suboptimal as it can lead to inefficient sampling and susceptibility to local minima. 
To address this, we design a gradient learning algorithm that incorporates global statistics measured by statistical algorithms like CMA-ES.
To that end, we introduce an importance weighting function W that assigns a higher weight to directions leading to lower values of the objective function. This encourages the algorithm to prioritize descent directions that not only follow the gradient but also align with globally favorable outcomes.

We define the Evolutionary Gradient Learning objective by adding an \textit{importance sampling} weight to the integral of Eq. \ref{eq:EGL_surrogate}
\begin{equation}
\label{eq:ogl_objective}
 g^{EvoGrad}_{\varepsilon}(x) = \operatorname*{arg\,min}_{g_{\theta}:\mathbb{R}^n \rightarrow \mathbb{R}^n} \int\limits_{\tau \in B_\varepsilon(0)} W(x + \tau) \cdot 
    \mathcal{R}_{g_\theta, x}^{EGL}(\tau)^2 d\tau
\end{equation}
The importance sampling factor \(W\) should be chosen s.t. it biases the optimization path towards lower regions regardless of the local curvature around \(x\), i.e. $W(x_1) \geq W(x_2)$ for \(f(x_1) \leq f(x_2)\). 

In our implementation, we train a CMA-ES with the samples generated by our uniform acquisition function around $x_k$ (where $k$ is the iteration index), and we use the evolved CMA distribution parameters as the integral weights, i.e.
\begin{equation}
W(x) \sim 
\exp\left( -\frac{1}{2\sigma^2} (x - m)^T C^{-1} (x - m) \right)
\end{equation}
Where $m,\sigma$ and $C$ are the CMA-ES evolving statistics as defined in Sec. \ref{sec:background}. This choice shifts the focus from the uniform acquisition function around $x_k$ to a higher quality area and biases the gradient towards the current global minimum inside the trust region.

\begin{figure}[ht]
    \centering
    \includegraphics[width=\columnwidth]{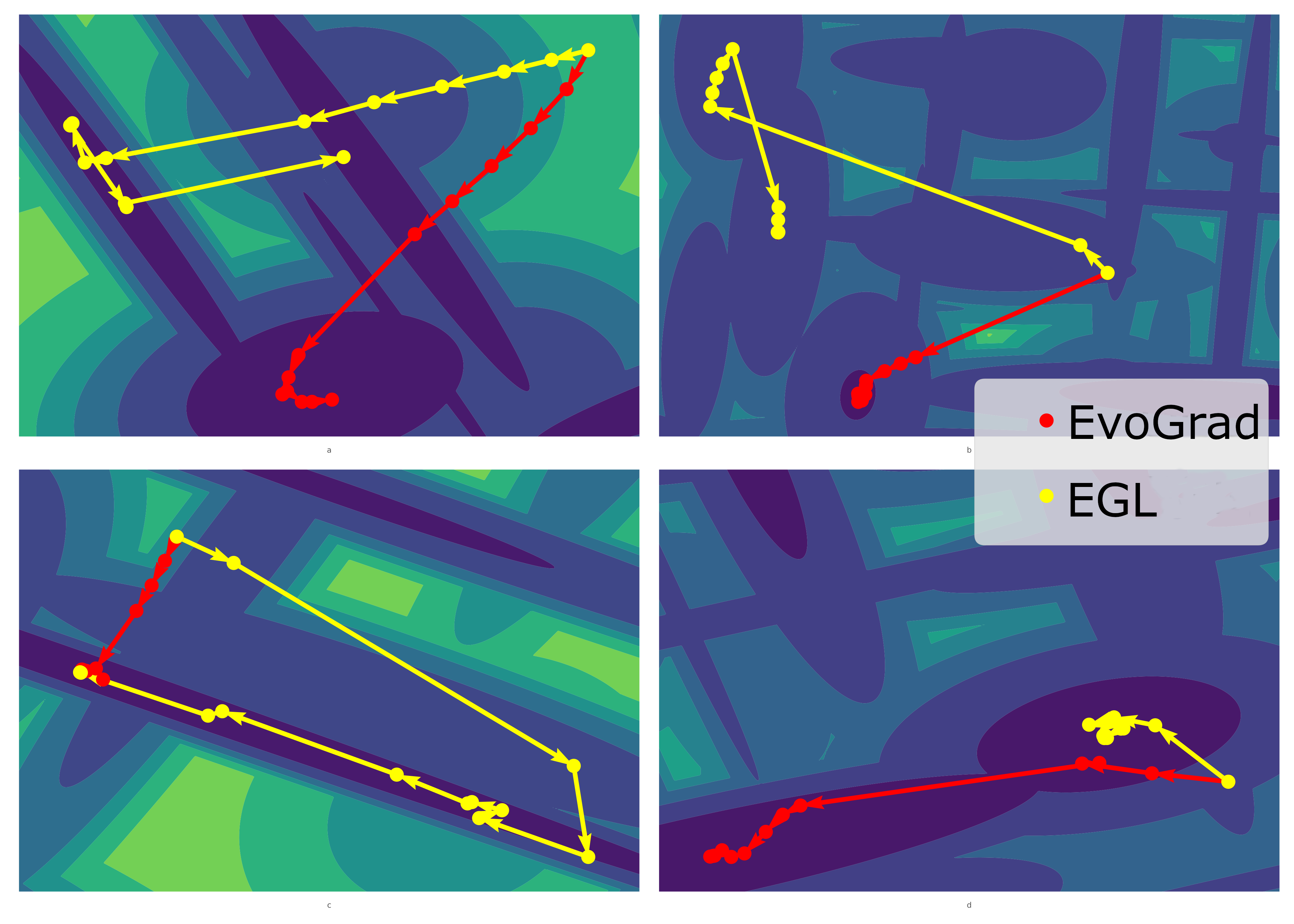}
    \caption{EvoGrad vs EGL trajectories. 1st row: Gallagher’s Gaussian 101-me. 2nd row: 21-hi.}
    \label{fig: compare_egl_and_weights}
\end{figure}

\begin{figure}[ht]
    \centering
    \includegraphics[width=\columnwidth]{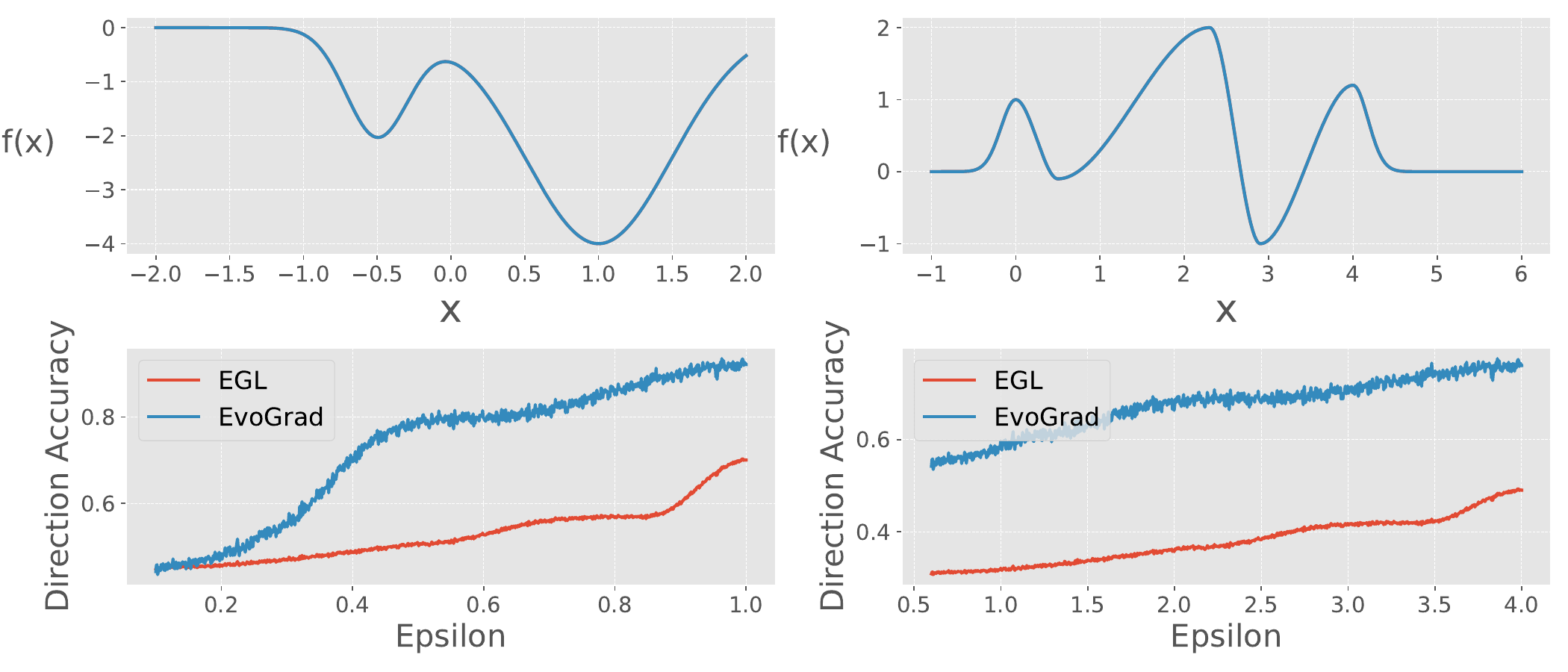}
    \caption{The probability for each algorithm to find the correct direction to the global minimum at a randomly selected starting point, based on the epsilon size.}
    \label{fig: close_to_min}
\end{figure}

Notice that in practice, the theoretical objective in Eq. \ref{eq:ogl_objective} is replaced by a sampled Monte-Carlo version (see Sec. \ref{sec:findings}) s.t. the sum of all weights across the sampled batch is smaller than 1. Additionally, to prevent the loss of gradient information during sampling, the function ensures that no sample is assigned a weight of exactly zero. Therefore, a minimum weight is enforced for each sample. This lower bound guarantees that all sampled directions contribute to the gradient estimate, and it also enables the proof for the \textit{controllable accuracy} (Appendix. \ref{eq:ogl_controllable_acc_proof}) property of EvoGrad, which implies that the mean-gradient converges to the true gradient, still holds for our biased version, s.t. when \(\varepsilon \to 0\), \(g^{EvoGrad}_{\varepsilon} \to \nabla f(x)\), this guarantees that the convergence properties of the mean-gradient still hold.
\begin{theorem} (Evolutionary Gradient Controllable Accuracy)
    For any differentiable function \(f\) with a continuous gradient, there exists \( \kappa^{EvoGrad} > 0 \) such that for any \(\varepsilon > 0\), \(g^{EvoGrad}_{\varepsilon}(x)\) satisfies
\[
\| g^{EvoGrad}_\varepsilon(x) - \nabla f(x) \| \leq \kappa^{EvoGrad} \varepsilon \quad \text{for all } x \in \Omega.
\]
\end{theorem}
To demonstrate the practical benefits of our evolutionary weighting approach, we plot 4 typical trajectories of EGL and EvoGrad in an environment with multiple local minima in Fig. \ref{fig: compare_egl_and_weights}. The trajectory of the weighted gradient is more direct, focusing on the global minimum, and is less easily distracted by nearby local minima. We also find that the biased version has a significantly higher probability of detecting the global minimum across different epsilon sizes, regardless of the starting point. In Fig. \ref{fig: close_to_min}, we show 2 1D functions and the probability for each algorithm to find the direction to the global minimum (relative to epsilon). For each value of $\varepsilon$, we sampled 500 random points, trained the gradient network \(g^{EvoGrad}_{\varepsilon}(x)\), and evaluated the resulting gradient direction. We can see that EvoGrad rapidly outperforms EGL, which is far more likely to get stuck in the local minimum, achieving substantially higher convergence rates to the global minimum, particularly as $\varepsilon$ increases.

\section{Gradient Learning with Hessian Corrections}
\label{sec:improvements}
To learn the mean-gradient, EGL minimizes the first-order Taylor residual (Sec. 2). Higher-order approximations can yield more accurate models, with the second-order Taylor expansion being
\begin{equation}
\label{eq:gradient_training_formula}
f(x + \tau) = f(x) + \nabla f(x)^\top \tau + \frac{1}{2} \tau^\top \nabla^2 f(x) \tau + R_2(x,x + \tau) \notag
\end{equation} 
Here \(R_2(x,y)=O(\|x-y\|^3)\) is the second order residual. By  replacing \(\nabla f\) with a surrogate model \(g_{\theta}\) and minimizing the resulting surrogate residual, we obtain our Higher-order Gradient Learning (HGrad) variant
\begin{align}
g^{HGrad}_{\varepsilon}(x) &= \operatorname*{arg\,min}_{g_{\theta}:\mathbb{R}^n \rightarrow \mathbb{R}^n} \int_{\tau \in B_\varepsilon(0)}\mathcal{R}_{g_\theta, x}^{HGrad}(\tau)^2d\tau \\
\mathcal{R}_{g_\theta, x}^{HGrad}(\tau) &= f(x) - f(x + \tau) + g_{\theta}(x)^\top \tau + \frac{1}{2}\tau^\top J_{g_{\theta}}(x) \tau \notag
\end{align}

The new higher-order term \(J_{g_{\theta}}(x)\) is the Jacobian of $g_{\theta}(x)$, evaluated at $x$ which approximates the function's Hessian matrix in the vicinity of our current solution, i.e., $J_{g_{\theta}}(x) \approx \nabla^2 f(x)$. Next, we show theoretically that, as expected, HGrad converges faster to the true gradient, which amounts to lower gradient error in practice.\footnote{Notice that, while HGrad incorporates Hessian corrections during the gradient learning phase, 
unlike  Newton’s methods, it is not used for scaling the gradient step size. Scaling the gradient requires calculation of the inverse Hessian, 
which is prone to numerical challenges and instabilities, and in our experiments was found less effective than the HGrad approach of minimizing the Taylor residual term.}
\begin{theorem} (Improved Controllable Accuracy):
For any twice differentiable function \( f \in \mathcal{C}^2 \), there exists \( \kappa_{HGrad} > 0 \) such that for any \( \varepsilon > 0 \), the second-order mean-gradient \( g^{HGrad}_{\varepsilon}(x) \) satisfies
\[
\| g^{HGrad}_{\varepsilon}(x) - \nabla f(x) \| \leq \kappa_{HGrad}\varepsilon^2 \quad \text{for all } x \in \Omega.
\]
\end{theorem}

In other words, in HGrad, the model's error is of an order of magnitude \(\varepsilon^2\) instead of \(\varepsilon\) in EGL and EvoGrad. 

Learning the gradient with the Jacobian corrections introduces a computational challenge, as double backpropagation can be expensive. This overhead can hinder the scalability and practical application of the method. A swift remedy is to detach the Jacobian matrix from the computation graph. While this step slightly changes the objective's gradient (i.e., the gradient through (\(\mathcal{R}^{HGrad}\)), it removes the second-order derivative, and in practice, we found that it achieves similar results compared to the full backpropagation through the residual \(\mathcal{R}^{HGrad}\), see Tab. \ref{tab:compare_metrics}.

\section{EvoGrad\textsuperscript{2} - Combining Evolutionary and Higher-Order Gradient Learning} 
\label{sec:findings}
Finally, we combine the Evolutionary Weighting method (Sec. \ref{sec:OGL}) with the Hessian corrections (Sec. \ref{sec:improvements}), presenting a practical implementation where integrals are replaced by Monte-Carlo sums over sampled pairs. The resulting loss function defines the Evolutionary Higher-order Gradient (EvoGrad\textsuperscript{2}) model
\begin{equation}
\label{eq:taylor_loss}
\mathcal{L}_{\varepsilon}^{EvoGrad\textsuperscript{2}}(\theta) = \sum_{\|x_i - x_j\| \leq \varepsilon} W(x_j)\mathcal{R}_{g_\theta, x_i}^{HGrad}(x_i - x_j)^2
\end{equation}
The summation is applied over sampled pairs which satisfy \(\|x_i - x_j\|\leq\varepsilon\). As explained in Sec. \ref{sec:improvements}, we detach the Jacobian of \(g_{\theta}\) from the computational graph to avoid second-order derivatives. Our final algorithm is outlined in the appendix (See Alg. \ref{code:full OHGL}).

\section{Experiments in the COCO test suite} \label{sec:experiment}

\begin{table*}[]
    \centering
    \scriptsize
    \setlength{\tabcolsep}{4pt}
    \renewcommand{\arraystretch}{1.2}
    
    \begin{tabularx}{\textwidth}{l|*{6}{>{\centering\arraybackslash}X|}>{\centering\arraybackslash}X}
        \textbf{Metric} 
        & \textbf{EvoGrad\textsuperscript{2}} & \textbf{EvoGrad} 
        & EvoGrad-0.1 & HGrad & HGrad-Attached 
        & EGL & IGL \\
        \hline
        Budget to solve
        & \textbf{44,712} & 51,859 
        & 54,972 & 61,147 & 61,221 
        & 58,488 & 75,031 \\
        Mean 
        & \textbf{0.003(0.02)} & 0.007(0.04) 
        & 0.072(0.1) & 0.014(0.07) & 0.013(0.07) 
        & 0.020(0.07) & 0.044(0.10) \\
        Solved Functions 
        & \textbf{0.949} & 0.917 
        & 0.71 & 0.858 & 0.858 
        & 0.821 & 0.674 \\
    \end{tabularx}

    \vspace{+2pt}

    \begin{tabularx}{\textwidth}{l|*{6}{>{\centering\arraybackslash}X|}>{\centering\arraybackslash}X}
        \textbf{Metric} 
        & CMA-TR & CMA & BFGS & SLSQP & Nelder-Mead & Powell &  \\
        \hline
        Budget to solve
        & 53,569 & 71,855 & 89,667 & 117,172 & 102,783 & 69,205 &  \\
        Mean 
        & 0.047(0.16) & 0.090(0.22)
        & 0.159(0.31) & 0.363(0.42) & 0.216(0.36) & 0.070(0.19) &  \\
        Solved Functions 
        & 0.740 & 0.612
        & 0.502 & 0.296 & 0.435 & 0.631 & \\
    \end{tabularx}

    \caption{Comparison of different metrics: Budget used to find value of 0.01 or lower ($\downarrow$), the mean normalized results ($\downarrow$), std ($\downarrow$); and percentage of solved problems ($\uparrow$).}
    \label{tab:compare_metrics}
\end{table*}

We evaluated the EvoGrad, HGrad, and EvoGrad\textsuperscript{2} algorithms on the COCO framework \cite{hansen2021coco} and compared them to EGL and other strong baselines: (1) CMA and its trust region variant CMA-TR, (2) Implicit Gradient Learning (IGL), where we train a model for the objective function and obtain the gradient estimation by backpropagation as in DDPG \cite{lillicrap2015continuous}, and (3) a variety of known algorithms (BFGS \cite{Nocedal2018NumericalO}, Nelder-Mead \cite{Nelder1965ASM}, Powell \cite{Powell1964AnEM}, SLSQP \cite{kraft1988software}), Using Scipy\footnote{https://scipy.org/} python package's implementation. We also adjusted EGL hyperparameters (See Appendix \ref{sec:sampling_size_adapt}) and improved the trust region (See Appendix \ref{sec:tr_management}) to reduce the budget usage. 

We use the following evaluation metrics:
\begin{itemize}
    \item \textbf{Performance}: The lowest objective value found within the given budget.
    \item \textbf{Success Rate}: Percentage of problems solved within a fixed budget.
    \item \textbf{Robustness}: Performance stability across different hyperparameter settings.
\end{itemize}

Performance was normalized against the best-known solutions to minimize bias: \(\texttt{normalized\_value} = \frac{y - y_{min}}{y_{max} - y_{min}}\). A function was considered solved if the normalized value was below 0.01.

\subsection{Success Rate and Performance}
\begin{figure*}[t]
    \centering
    \includegraphics[width=1\textwidth]{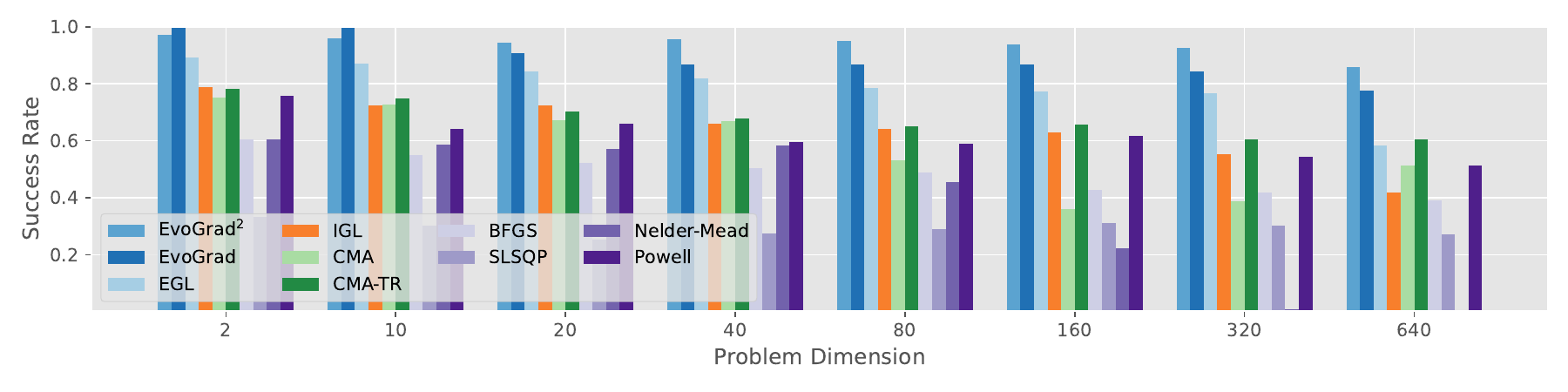}
    \caption{The probability of each algorithm to solve a problem in each dimension }
    \label{fig:solved_by_dim}
\end{figure*}

For the first experiment, observe Figure \ref{fig:solved_by_dim}, where we compare the success rates of the benchmark algorithms by dimension. The dimensional analysis shows that as the dimension grows, traditional optimization methods suffer. While the trust region was able to improve the results of the CMA variants, it still has a drop in performance in the dimensions above 40.
This performance degradation reflects the fundamental challenge facing sample-based BBO methods: the curse of dimensionality demands exponentially more evaluations as problem size grows. Traditional approaches that work effectively in lower-dimensional spaces become increasingly sample-inefficient in higher dimensions.
On the other hand, using gradient learning, we found multiple methods to reduce the samples required for an effective search, as EvoGrad and EvoGrad\textsuperscript{2} both maintain a steady success rate across dimensions. 

\subsection{Budget-Constrained Applications}
In our second experiment, we analyze how different algorithms perform under varying application constraints to understand which method is most suitable for specific use cases. Figure \ref{fig:experiements_results} presents an analysis across critical considerations that directly impact real-world deployment decisions.
Fig. \ref{fig:experiements_results}(a) illustrates algorithm performance under different budget constraints. While the gradient-based method has a slower start (Due to warm-up training, which consumes early budget), both EvoGrad and EvoGrad\textsuperscript{2} quickly pick up the pace, finally outperforming the competitors around 8K samples. While CMA has an advantage in the low-budget regime, our algorithm shines when high-accuracy solution is needed, as shown in Table \ref{tab:compare_metrics}, EvoGrad and  EvoGrad\textsuperscript{2} have a clear advantage for solving the function in a minimum budget.    
For low-budget applications where function evaluations are expensive, the results show that while CMA variants are initially efficient at improving performance, they plateau earlier. In contrast, EvoGrad and EvoGrad\textsuperscript{2} demonstrate superior long-term performance, making them ideal for applications where the total budget allows for extended optimization runs.

\subsection{High-Accuracy Requirements}
\begin{figure*}[h]
    \centering
    \includegraphics[width=1\textwidth]{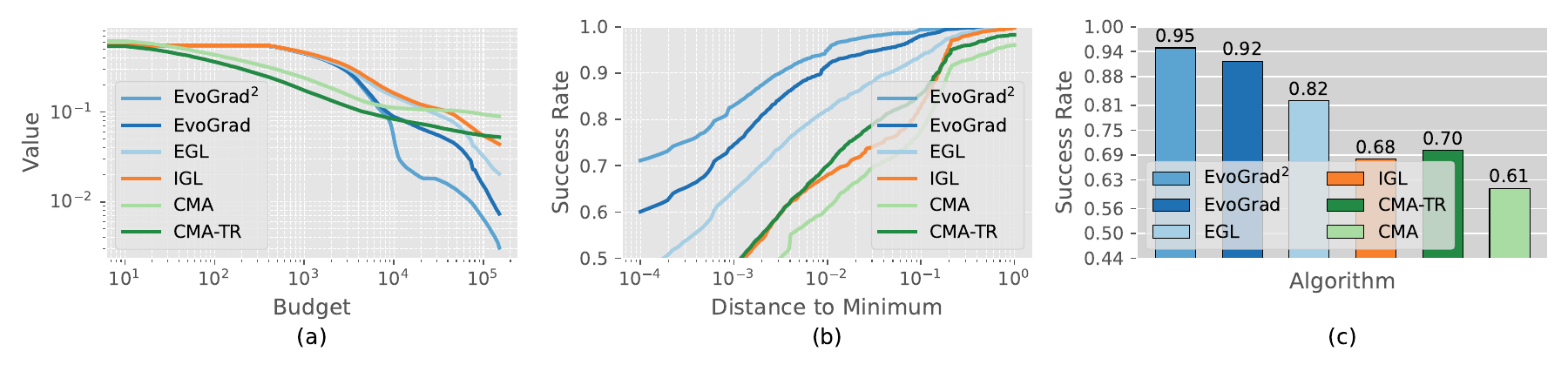}
    \caption{Experiment results against the baseline: (a) Convergence for all our algorithms against baseline algorithms, (b) Success rate as a function of the normalized distance from the best-known solution, (c) Percentage of solved algorithms when the distance from the best point is 0.01}
    \label{fig:experiements_results}
\end{figure*}

Figure \ref{fig:experiements_results}(b) examines success rates as a function of distance to the optimal solution, which is crucial for applications requiring high precision. The results demonstrate that EvoGrad and EvoGrad\textsuperscript{2} consistently maintain higher success rates across all accuracy thresholds. This is particularly important for applications such as adversarial attack generation or hyperparameter tuning, where finding near-optimal solutions is critical.

Finally, Fig. \ref{fig:experiements_results}(c) and our t-test analysis in Table \ref{tab:t_test_comparison} confirm that our gains are statistically significant. For every baseline, the p-value is below $10^{-6}$, indicating that both EvoGrad and EvoGrad\textsuperscript{2} outperform alternative methods well beyond the $\alpha$ = 0.05 threshold.

\subsection{Ablation}
For the EvoGrad algorithm (Sec. \ref{sec:findings}), we explored different weighting schemes for biased sampling. Our final approach uses CMA's covariance matrix to create the importance sampler, which adapts to the local landscape geometry. We compared this against simpler alternatives: EvoGrad-0.1 uses the softmax function to create $W$. Assigning probabilities for each sample $x \in \mathcal{D}_k$ based on $f(x)$. Table 1 shows that the adaptive covariance matrix significantly outperforms simpler weighting schemes, particularly in higher dimensions.

For HGrad (Sec. \ref{sec:improvements}), computing second-order derivatives can be memory-intensive. We found that detaching the Jacobian from the computational graph (HGrad vs HGrad-Attached in Table 1) maintains nearly identical performance while reducing memory overhead, making the approach more scalable.

\subsection{Hyperparameter Tolerance}
We evaluate the robustness of our algorithm to hyperparameter tuning. Our objective is to show which hyperparameters have an impact on the algorithm's performance and how to set those parameters. We conducted systematic experiments to assess this, modifying several hyperparameters and analyzing their effects. Table \ref{tab:mean_std_group} (and Table \ref{tab:algorithm_versions}
in the appendix) reports the coefficient of variation (CV), defined as $CV = \frac{\sigma}{\mu}$, across different hyperparameter sweeps, highlighting the algorithm's stability under varying conditions.


Our findings show that hyperparameters such as the epsilon factor, shrink factor, and learning rate (LR) have cumulative effects: small variations may seem negligible but can significantly impact performance over time. In \ref{eq:hgl_convergence_proof}, we establish the step size–epsilon relationship needed for progress, while the shrink factor should align with the budget to maximize explored sub-problems. Thus, fine-tuning these parameters is critical. By contrast, network structure (layers and size) had little effect, suggesting that the Taylor loss enables effective learning even with simple architectures, and greater complexity does not necessarily improve results.

\begin{table}[h]
    \centering
    \scriptsize
    \resizebox{\columnwidth}{!}{%
    \begin{tabular}{l|c|c|c|c|c|c}
        \textbf{Metric} & \textbf{Networks} & \textbf{$\varepsilon$} & \textbf{$\varepsilon$-Factor} & \textbf{TR SF} & \textbf{LR} & \textbf{Gradient LR} \\
        \hline
        CV & 0.0429  & 0.013  & 0.1676  & 0.1691  & 0.2328 & 0.1263 \\
    \end{tabular}%
    }
    \caption{Coefficient Variation ($CV=\frac{\sigma}{\mu}$) over Hyperparameter sweep experiment. TR SF = trust region shrink factor.}
    \label{tab:mean_std_group}
\end{table}

\section{High Dimensional Applications}
\label{sec:applications}
\subsection{Adversarial Attacks}

As powerful vision models like ResNet \cite{targ2016resnet} and Vision Transformers (ViT) \cite{han2022survey} grow in prominence, adversarial attacks have become a significant concern. These attacks subtly modify inputs, causing models to misclassify them, while the perturbation remains imperceptible to both human vision and other classifiers \cite{tang2019adversarial}.
We can define it formally:

\begin{equation}
\label{eq:advarserial_min}
    x^\ast_a = \arg\min_{x} d(x, x_a) \ \ \ \text{s.t.} f(x) \neq f(x_a)
\end{equation}
Where $f$ is the classifier and $d$ is some distance metric between elements.

Recent studies have extended adversarial attacks to domains like AI-text detection \cite{sadasivan2024aigeneratedtextreliablydetected} and automotive sensors \cite{Mahima3DAttackReview}. These attacks prevent tracking and detection, posing risks to both users and pedestrians.
Adversarial attacks are classified into black-box and white-box methods. Black-box attacks only require query access, while white-box methods use model gradients to craft perturbations \cite{machado2021adversarial,cao2019adversarial}. Despite some black-box methods relying on surrogate models \cite{dong2018boosting,xiao2018generating,madry2017towards,goodfellow2014explaining}, approaches like~\cite{tu2019autozoom} generate random samples to approximate gradient estimation, though they are computationally expensive. Other methods use GAN networks to search latent spaces for adversarial examples \cite{liu2021multi,sarkar2017upset}. Still, they depend on existing GANs and their latent space diversity.

Our EvoGrad\textsuperscript{2} method offers a true black-box approach with precise perturbation control, avoiding gradient back-propagation. EvoGrad\textsuperscript{2} directly optimizes perturbations to maintain low distortion while fooling the model, handling high-dimensional spaces with over 30K parameters.

\begin{table}[h]
    \centering
    \scriptsize 
    \resizebox{\columnwidth}{!}{%
    \begin{tabular}{l|c|c|c|c|c}
        \textbf{Metric} & \textbf{CMA} & \textbf{EvoGrad} & \textbf{HGrad} & \textbf{EvoGrad\textsuperscript{2}} & \textbf{CMA+EvoGrad} \\
        \hline
        Accuracy & 0.02 & 0.02 & 0.02 & 0.02 & 0.02 \\
        MSE & 0.05 & \textbf{0.001} & 0.002 & \textbf{0.001} & 0.001\\
        Time (Until Convergence) & \textbf{20m} & 6H & 7H & 7H & 3H \\
    \end{tabular}%
    }
    \caption{Methods' comparison on Accuracy, MSE, Time.}
    \label{tab:compare_attack_methods}
\end{table}

\begin{figure}[h]
    \centering
    \begin{minipage}[b]{0.1\textwidth}
        \centering
        \includegraphics[width=\textwidth]{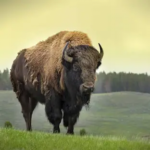}
        \subcaption{Original}
    \end{minipage}
    \begin{minipage}[b]{0.1\textwidth}
        \centering
        \includegraphics[width=\textwidth]{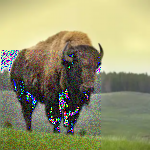}
        \subcaption{CMA}
    \end{minipage}
    \begin{minipage}[b]{0.1\textwidth}
        \centering
        \includegraphics[width=\textwidth]{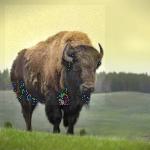}
        \subcaption{EvoGrad}
    \end{minipage}
    \begin{minipage}[b]{0.1\textwidth}
        \centering
        \includegraphics[width=\textwidth]{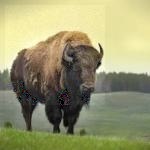}
        \subcaption{EvoGrad\textsuperscript{2}}
    \end{minipage}
    \caption{Adversarial examples generated by EvoGrad and CMA  against the ImageNet model.}
    \label{fig:set1}
\end{figure}

\textbf{Methodology}: We tested our algorithms against classifiers trained on MNIST, CIFAR-10, and ImageNet, aiming to minimize Eq. \eqref{eq:advarserial_min}. To generate adversarial images with minimal distortion, we developed a penalty that jointly minimizes MSE and CE-loss (Appendix. \ref{sec:advarserial_atack_implementation}). This approach successfully fooled the model, evading the top 5 classifications.

\textbf{Results}: We evaluated five different configurations: CMA, EvoGrad, HGrad, EvoGrad\textsuperscript{2}, and a combination of both (CMA+EvoGrad), where the CMA run provides the initial guess for an EvoGrad run. While CMA alone was not able to converge to a satisfying adversarial example, the combined CMA+EvoGrad enjoyed the rapid start of CMA with the robustness of EvoGrad, s.t. it was able to find a satisfying adversarial example half the computation time of EvoGrad and EvoGrad\textsuperscript{2}.

\subsection{Code Generation}
The development of large language models (LLMs) such as Transformers \cite{vaswani2017attention} have advanced code generation \cite{dehaerne2022code}. Despite these strides, fine-tuning outputs based on parameters measured post-generation remains challenging.
Recent algorithms have been developed to generate code tailored for specific tasks using LLMs. For instance, FunSearch \cite{romera2024mathematical} generates new code solutions for complex tasks, while Chain of Code \cite{li2023chain} incorporates reasoning to detect and correct errors in the output code. Similarly, our method uses black-box optimization to guide code generation for runtime efficiency.
Building on \cite{zhang2024interpreting}, which links LLM expertise to a small parameter set, we fine-tuned the embedding layer to reduce Python code runtime. Using LoRA \cite{lora}, we optimized the generated code based on execution time, scaling up to $\sim 200\text{k}$ parameters.

\textbf{Fibonacci:}
We tested this approach by having the model generate a Fibonacci function. Initial results were incorrect, optimization guided the model to a correct and efficient solution. Figure \ref{fig:code_gen_sample1} illustrates this progression, with the 25th step showing an optimized version.

\textbf{Line-Level Efficiency Enhancements:}
We tested the model's ability to implement small code efficiencies, such as replacing traditional for-loops with list comprehensions. The algorithm optimized the order of four functions—`initialize`, `start`, `activate`, and `stop`—each with eight variants, minimizing overall runtime by optimizing the function order.

\textbf{Code Force:} For a more complex problem, we used the Count Triplets challenge from Codeforces(\url{https://codeforces.com/}). While the model initially struggled, once it found a correct solution, the algorithm further optimized it for runtime performance (See Appendix. \ref{sec:code_force_generated_code}).

\begin{figure}[t]
    \centering
    \includegraphics[width=\columnwidth]{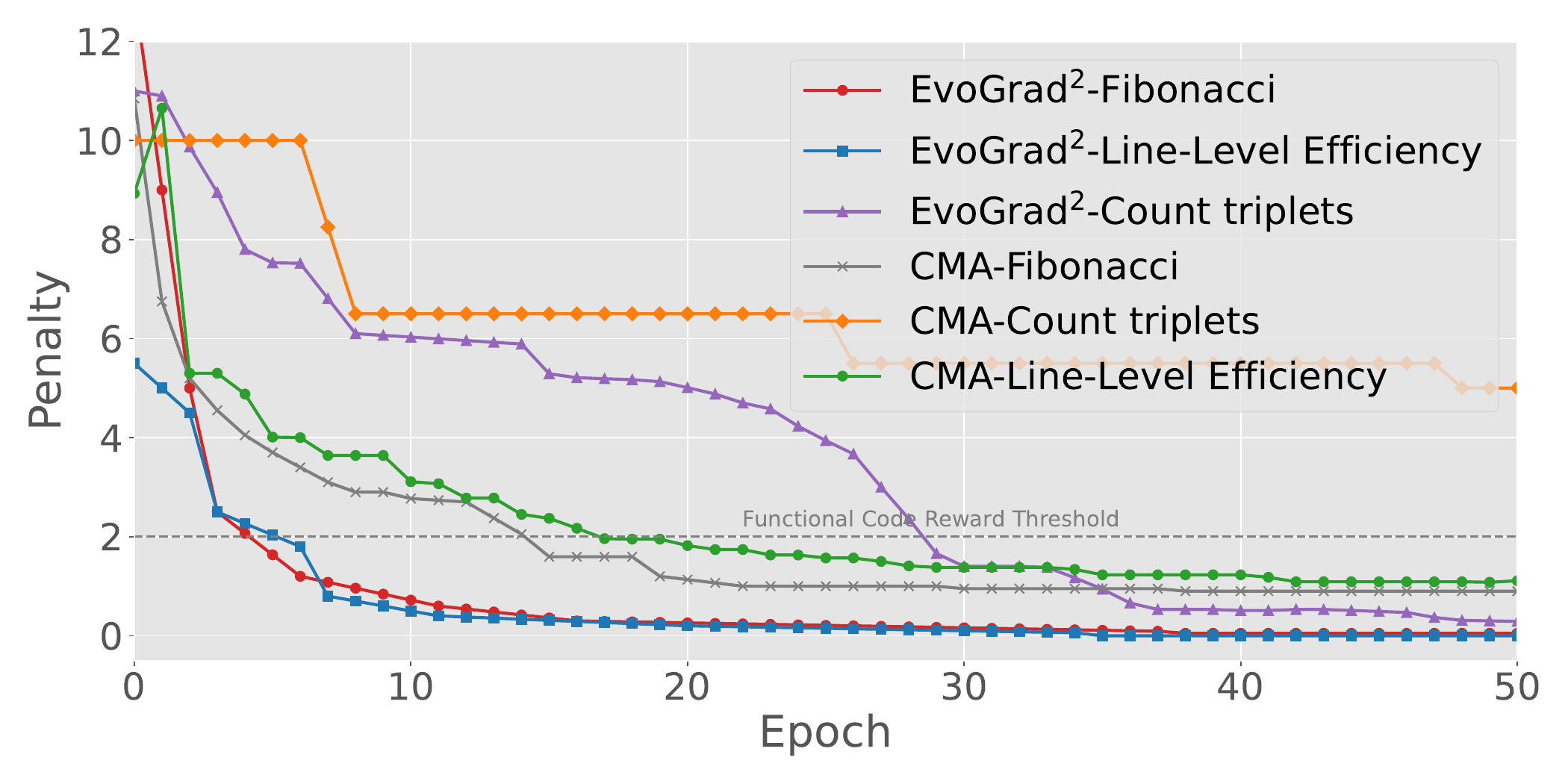}
    \caption{Code generation experiments: penalty over time.}
    \label{fig:code_opt_convergence}
\end{figure}

\textbf{Discussion:} Our method demonstrates the ability to generate correct solutions while applying micro-optimizations for efficiency. In simple tasks like Fibonacci, EvoGrad\textsuperscript{2} converged on an optimal solution (See Fig. \ref{fig:code_opt_convergence}), and in more complex problems, it improved the initial solutions. However, in more complex tasks, the LLM may generate code that fails to solve the problem, hindering the optimization of the run-time. To address this, either stronger models or methods focused on optimizing solution correctness are needed. This would ensure that valid solutions are generated first, which can then be further optimized for performance.
\section{Conclusion} \label{sec:conclusion}


We introduced EvoGrad\textsuperscript{2}, a novel black-box optimizer that integrates local gradient learning with global evolutionary statistics, making it well-suited for high-dimensional, non-convex problems. By incorporating second-order Taylor approximations with Hessian corrections, it improves gradient estimation, and to our knowledge, is the first method to combine neural gradient learning with evolutionary algorithms.

We believe that future research should extend EvoGrad\textsuperscript{2} to contemporary core machine-learning problems. For example, optimize Large Language Models in Reinforcement Learning with Human Feedback (RLHF) and Verifiable Rewards (RLVR)\cite{bai2022training,guo2025deepseek}, enabling more data-efficient, off-policy learning.
Additionally, exploring EvoGrad\textsuperscript{2}'s integration into diffusion, score-based, and flow-matching models \cite{song2019generative,song2020score,lipman2022flow} could reduce computational overhead in iterative gradient evaluations and expand its applicability in generative modeling.

\newpage

\bigskip

\bibliography{aaai2026}

@article{alarie2021two,
  title={Two decades of blackbox optimization applications},
  author={Alarie, St{\'e}phane and Audet, Charles and Gheribi, A{\"\i}men E and Kokkolaras, Michael and Le Digabel, S{\'e}bastien},
  journal={EURO Journal on Computational Optimization},
  volume={9},
  pages={100011},
  year={2021},
  publisher={Elsevier}
}

@article{zhen2014simulation,
  title={A simulation optimization framework for ambulance deployment and relocation problems},
  author={Zhen, Lu and Wang, Kai and Hu, Hongtao and Chang, Daofang},
  journal={Computers \& Industrial Engineering},
  volume={72},
  pages={12--23},
  year={2014},
  publisher={Elsevier}
}

@inproceedings{gehring2014towards,
  title={Towards automatic discovery of agile gaits for quadrupedal robots},
  author={Gehring, Christian and Coros, Stelian and Hutter, Marco and Bloesch, Michael and Fankhauser, P{\'e}ter and Hoepflinger, Markus A and Siegwart, Roland},
  booktitle={2014 IEEE international conference on robotics and automation (ICRA)},
  pages={4243--4248},
  year={2014},
  organization={IEEE}
}

@article{ding2024longrope,
  title={Longrope: Extending llm context window beyond 2 million tokens},
  author={Ding, Yiran and Zhang, Li Lyna and Zhang, Chengruidong and Xu, Yuanyuan and Shang, Ning and Xu, Jiahang and Yang, Fan and Yang, Mao},
  journal={arXiv preprint arXiv:2402.13753},
  year={2024}
}

@article{guo2025deepseek,
  title={Deepseek-r1: Incentivizing reasoning capability in llms via reinforcement learning},
  author={Guo, Daya and Yang, Dejian and Zhang, Haowei and Song, Junxiao and Zhang, Ruoyu and Xu, Runxin and Zhu, Qihao and Ma, Shirong and Wang, Peiyi and Bi, Xiao and others},
  journal={arXiv preprint arXiv:2501.12948},
  year={2025}
}

@article{song2020score,
  title={Score-based generative modeling through stochastic differential equations},
  author={Song, Yang and Sohl-Dickstein, Jascha and Kingma, Diederik P and Kumar, Abhishek and Ermon, Stefano and Poole, Ben},
  journal={arXiv preprint arXiv:2011.13456},
  year={2020}
}

@article{lipman2022flow,
  title={Flow matching for generative modeling},
  author={Lipman, Yaron and Chen, Ricky TQ and Ben-Hamu, Heli and Nickel, Maximilian and Le, Matt},
  journal={arXiv preprint arXiv:2210.02747},
  year={2022}
}

@article{song2019generative,
  title={Generative modeling by estimating gradients of the data distribution},
  author={Song, Yang and Ermon, Stefano},
  journal={Advances in neural information processing systems},
  volume={32},
  year={2019}
}

@article{prabhu2018survey,
  title={A survey on evolutionary-aided design in robotics},
  author={Prabhu, Shanker G Radhakrishna and Seals, Richard C and Kyberd, Peter J and Wetherall, Jodie C},
  journal={Robotica},
  volume={36},
  number={12},
  pages={1804--1821},
  year={2018},
  publisher={Cambridge University Press}
}

@misc{olof2018comparative,
  title={A comparative study of black-box optimization algorithms for tuning of hyper-parameters in deep neural networks},
  author={Olof, Skogby Steinholtz},
  year={2018}
}

@article{audet2017introduction,
  title={Chapter 1: The begining of DFO algorithms in derivative-free and blackbox optimization},
  author={Audet, Charles and Hare, Warren and Audet, Charles and Hare, Warren},
  journal={Derivative-Free and Blackbox Optimization},
  pages={11--15},
  year={2017},
  publisher={Springer}
}

@article{bischl2023hyperparameter,
  title={Hyperparameter optimization: Foundations, algorithms, best practices, and open challenges},
  author={Bischl, Bernd and Binder, Martin and Lang, Michel and Pielok, Tobias and Richter, Jakob and Coors, Stefan and Thomas, Janek and Ullmann, Theresa and Becker, Marc and Boulesteix, Anne-Laure and others},
  journal={Wiley Interdisciplinary Reviews: Data Mining and Knowledge Discovery},
  volume={13},
  number={2},
  pages={e1484},
  year={2023},
  publisher={Wiley Online Library}
}

@inproceedings{bouneffouf2020survey,
  title={Survey on applications of multi-armed and contextual bandits},
  author={Bouneffouf, Djallel and Rish, Irina and Aggarwal, Charu},
  booktitle={2020 IEEE Congress on Evolutionary Computation (CEC)},
  pages={1--8},
  year={2020},
  organization={IEEE}
}

@article{bai2022training,
  title={Training a helpful and harmless assistant with reinforcement learning from human feedback},
  author={Bai, Yuntao and Jones, Andy and Ndousse, Kamal and Askell, Amanda and Chen, Anna and DasSarma, Nova and Drain, Dawn and Fort, Stanislav and Ganguli, Deep and Henighan, Tom and others},
  journal={arXiv preprint arXiv:2204.05862},
  year={2022}
}

@article{anil2020scalable,
  title={Scalable second order optimization for deep learning},
  author={Anil, Rohan and Gupta, Vineet and Koren, Tomer and Regan, Kevin and Singer, Yoram},
  journal={arXiv preprint arXiv:2002.09018},
  year={2020}
}

@article{audet2017searchmethods,
  title={Chapter 3: The begining of DFO algorithms in derivative-free and blackbox optimization},
  author={Audet, Charles and Hare, Warren and Audet, Charles and Hare, Warren},
  journal={Derivative-Free and Blackbox Optimization},
  pages={33--47},
  year={2017},
  publisher={Springer}
}

@article{audet2017genetic,
  title={Genetic methods in derivative-free and blackbox optimization},
  author={Audet, Charles and Hare, Warren and Audet, Charles and Hare, Warren},
  journal={Derivative-Free and Blackbox Optimization},
  pages={57--73},
  year={2017},
  publisher={Springer}
}

@article{audet2017model,
  title={Model-based methods in derivative-free and blackbox optimization},
  author={Audet, Charles and Hare, Warren and Audet, Charles and Hare, Warren},
  journal={Derivative-Free and Blackbox Optimization},
  pages={156--218},
  year={2017},
  publisher={Springer}
}

@article{audet2017direct,
  title={DIRECT search in derivative-free and blackbox optimization},
  author={Audet, Charles and Hare, Warren and Audet, Charles and Hare, Warren},
  journal={Derivative-Free and Blackbox Optimization},
  pages={93--156},
  year={2017},
  publisher={Springer}
}

@article{busetti2003simulated,
  title={Simulated annealing overview},
  author={Busetti, Franco},
  journal={World Wide Web URL www. geocities. com/francorbusetti/saweb. pdf},
  volume={4},
  year={2003}
}

@article{hansen2016cma,
  title={The CMA evolution strategy: A tutorial},
  author={Hansen, Nikolaus},
  journal={arXiv preprint arXiv:1604.00772},
  year={2016}
}

@book{clerc2010particle,
  title={Particle swarm optimization},
  author={Clerc, Maurice},
  volume={93},
  year={2010},
  publisher={John Wiley \& Sons}
}

@inproceedings{loshchilov2013bi,
  title={Bi-population CMA-ES agorithms with surrogate models and line searches},
  author={Loshchilov, Ilya and Schoenauer, Marc and Sebag, Mich{\`e}le},
  booktitle={Proceedings of the 15th annual conference companion on Genetic and evolutionary computation},
  pages={1177--1184},
  year={2013}
}

@inproceedings{sarafian2020explicit,
  title={Explicit Gradient Learning for Black-Box Optimization.},
  author={Sarafian, Elad and Sinay, Mor and Louzoun, Yoram and Agmon, Noa and Kraus, Sarit},
  booktitle={ICML},
  pages={8480--8490},
  year={2020}
}

@article{rimon2024mamba,
  title={Mamba: an effective world model approach for meta-reinforcement learning},
  author={Rimon, Zohar and Jurgenson, Tom and Krupnik, Orr and Adler, Gilad and Tamar, Aviv},
  journal={arXiv preprint arXiv:2403.09859},
  year={2024}
}

@ARTICLE{hansen2021coco,
author = {Hansen, N. and Auger, A. and Ros, R. and Mersmann, O.
          and Tu{\v s}ar, T. and Brockhoff, D.},
title = {{COCO}: A Platform for Comparing Continuous Optimizers 
          in a Black-Box Setting},
journal = {Optimization Methods and Software},
doi = {https://doi.org/10.1080/10556788.2020.1808977},
pages = {114--144},
issue = {1},
volume = {36},
year = 2021
}

@inproceedings{hansen2010comparing,
  title={Comparing results of 31 algorithms from the black-box optimization benchmarking BBOB-2009},
  author={Hansen, Nikolaus and Auger, Anne and Ros, Raymond and Finck, Steffen and Po{\v{s}}{\'\i}k, Petr},
  booktitle={Proceedings of the 12th annual conference companion on Genetic and evolutionary computation},
  pages={1689--1696},
  year={2010}
}

@book{back1996evolutionary,
  title={Evolutionary algorithms in theory and practice: evolution strategies, evolutionary programming, genetic algorithms},
  author={Back, Thomas},
  year={1996},
  publisher={Oxford university press}
}

@book{bertsekas2015convex,
  title={Convex optimization algorithms},
  author={Bertsekas, Dimitri},
  year={2015},
  publisher={Athena Scientific}
}

@article{targ2016resnet,
  title={Resnet in resnet: Generalizing residual architectures},
  author={Targ, Sasha and Almeida, Diogo and Lyman, Kevin},
  journal={arXiv preprint arXiv:1603.08029},
  year={2016}
}

@article{han2022survey,
  title={A survey on vision transformer},
  author={Han, Kai and Wang, Yunhe and Chen, Hanting and Chen, Xinghao and Guo, Jianyuan and Liu, Zhenhua and Tang, Yehui and Xiao, An and Xu, Chunjing and Xu, Yixing and others},
  journal={IEEE transactions on pattern analysis and machine intelligence},
  volume={45},
  number={1},
  pages={87--110},
  year={2022},
  publisher={IEEE}
}

@article{tang2019adversarial,
  title={Adversarial attack type I: Cheat classifiers by significant changes},
  author={Tang, Sanli and Huang, Xiaolin and Chen, Mingjian and Sun, Chengjin and Yang, Jie},
  journal={IEEE transactions on pattern analysis and machine intelligence},
  volume={43},
  number={3},
  pages={1100--1109},
  year={2019},
  publisher={IEEE}
}

@article{machado2021adversarial,
  title={Adversarial machine learning in image classification: A survey toward the defender’s perspective},
  author={Machado, Gabriel Resende and Silva, Eug{\^e}nio and Goldschmidt, Ronaldo Ribeiro},
  journal={ACM Computing Surveys (CSUR)},
  volume={55},
  number={1},
  pages={1--38},
  year={2021},
  publisher={ACM New York, NY}
}

@inproceedings{dong2018boosting,
  title={Boosting adversarial attacks with momentum},
  author={Dong, Yinpeng and Liao, Fangzhou and Pang, Tianyu and Su, Hang and Zhu, Jun and Hu, Xiaolin and Li, Jianguo},
  booktitle={Proceedings of the IEEE conference on computer vision and pattern recognition},
  pages={9185--9193},
  year={2018}
}

@article{xiao2018generating,
  title={Generating adversarial examples with adversarial networks},
  author={Xiao, Chaowei and Li, Bo and Zhu, Jun-Yan and He, Warren and Liu, Mingyan and Song, Dawn},
  journal={arXiv preprint arXiv:1801.02610},
  year={2018}
}

@article{madry2017towards,
  title={Towards deep learning models resistant to adversarial attacks},
  author={Madry, Aleksander and Makelov, Aleksandar and Schmidt, Ludwig and Tsipras, Dimitris and Vladu, Adrian},
  journal={arXiv preprint arXiv:1706.06083},
  year={2017}
}

@inproceedings{tu2019autozoom,
  title={Autozoom: Autoencoder-based zeroth order optimization method for attacking black-box neural networks},
  author={Tu, Chun-Chen and Ting, Paishun and Chen, Pin-Yu and Liu, Sijia and Zhang, Huan and Yi, Jinfeng and Hsieh, Cho-Jui and Cheng, Shin-Ming},
  booktitle={Proceedings of the AAAI conference on artificial intelligence},
  volume={33},
  number={01},
  pages={742--749},
  year={2019}
}

@article{vaswani2017attention,
  title={Attention is all you need},
  author={Vaswani, A},
  journal={Advances in Neural Information Processing Systems},
  year={2017}
}

@article{lora,
  title={Lora: Low-rank adaptation of large language models},
  author={Hu, Edward J and Shen, Yelong and Wallis, Phillip and Allen-Zhu, Zeyuan and Li, Yuanzhi and Wang, Shean and Wang, Lu and Chen, Weizhu},
  journal={arXiv preprint arXiv:2106.09685},
  year={2021}
}

@inproceedings{tang2021guiding,
  title={Guiding Evolutionary Strategies with Off-Policy Actor-Critic.},
  author={Tang, Yunhao},
  booktitle={AAMAS},
  pages={1317--1325},
  year={2021}
}

@article{romera2024mathematical,
  title={Mathematical discoveries from program search with large language models},
  author={Romera-Paredes, Bernardino and Barekatain, Mohammadamin and Novikov, Alexander and Balog, Matej and Kumar, M Pawan and Dupont, Emilien and Ruiz, Francisco JR and Ellenberg, Jordan S and Wang, Pengming and Fawzi, Omar and others},
  journal={Nature},
  volume={625},
  number={7995},
  pages={468--475},
  year={2024},
  publisher={Nature Publishing Group UK London}
}

@article{zhang2024interpreting,
  title={Interpreting and Improving Large Language Models in Arithmetic Calculation},
  author={Zhang, Wei and Wan, Chaoqun and Zhang, Yonggang and Cheung, Yiu-ming and Tian, Xinmei and Shen, Xu and Ye, Jieping},
  journal={arXiv preprint arXiv:2409.01659},
  year={2024}
}

@article{li2023chain,
  title={Chain of code: Reasoning with a language model-augmented code emulator},
  author={Li, Chengshu and Liang, Jacky and Zeng, Andy and Chen, Xinyun and Hausman, Karol and Sadigh, Dorsa and Levine, Sergey and Fei-Fei, Li and Xia, Fei and Ichter, Brian},
  journal={arXiv preprint arXiv:2312.04474},
  year={2023}
}

@article{goodfellow2014explaining,
  title={Explaining and harnessing adversarial examples},
  author={Goodfellow, Ian J and Shlens, Jonathon and Szegedy, Christian},
  journal={arXiv preprint arXiv:1412.6572},
  year={2014}
}

@misc{sadasivan2024aigeneratedtextreliablydetected,
      title={Can AI-Generated Text be Reliably Detected?}, 
      author={Vinu Sankar Sadasivan and Aounon Kumar and Sriram Balasubramanian and Wenxiao Wang and Soheil Feizi},
      year={2024},
      eprint={2303.11156},
      archivePrefix={arXiv},
      primaryClass={cs.CL},
      url={https://arxiv.org/abs/2303.11156}, 
}

@inproceedings{liu2021multi,
  title={Multi-view correlation based black-box adversarial attack for 3D object detection},
  author={Liu, Bingyu and Guo, Yuhong and Jiang, Jianan and Tang, Jian and Deng, Weihong},
  booktitle={Proceedings of the 27th ACM SIGKDD Conference on Knowledge Discovery \& Data Mining},
  pages={1036--1044},
  year={2021}
}

@article{sarkar2017upset,
  title={UPSET and ANGRI: Breaking high performance image classifiers},
  author={Sarkar, Sayantan and Bansal, Ankan and Mahbub, Upal and Chellappa, Rama},
  journal={arXiv preprint arXiv:1707.01159},
  year={2017}
}

@ARTICLE{Mahima3DAttackReview,
  author={Mahima, K. T. Yasas and Perera, Asanka G. and Anavatti, Sreenatha and Garratt, Matt},
  journal={IEEE Transactions on Intelligent Transportation Systems}, 
  title={Toward Robust 3D Perception for Autonomous Vehicles: A Review of Adversarial Attacks and Countermeasures}, 
  year={2024},
  volume={},
  number={},
  pages={1-27},
  doi={10.1109/TITS.2024.3456293}}

@article{cao2019adversarial,
  title={Adversarial objects against lidar-based autonomous driving systems},
  author={Cao, Yulong and Xiao, Chaowei and Yang, Dawei and Fang, Jing and Yang, Ruigang and Liu, Mingyan and Li, Bo},
  journal={arXiv preprint arXiv:1907.05418},
  year={2019}
}

@ARTICLE{zoReview,
  author={Liu, Sijia and Chen, Pin-Yu and Kailkhura, Bhavya and Zhang, Gaoyuan and Hero III, Alfred O. and Varshney, Pramod K.},
  journal={IEEE Signal Processing Magazine}, 
  title={A Primer on Zeroth-Order Optimization in Signal Processing and Machine Learning: Principals, Recent Advances, and Applications}, 
  year={2020},
  volume={37},
  number={5},
  pages={43-54},
  doi={10.1109/MSP.2020.3003837}}

@article{dehaerne2022code,
  title={Code generation using machine learning: A systematic review},
  author={Dehaerne, Enrique and Dey, Bappaditya and Halder, Sandip and De Gendt, Stefan and Meert, Wannes},
  journal={Ieee Access},
  volume={10},
  pages={82434--82455},
  year={2022},
  publisher={IEEE}
}

@article{golovin2019gradientless,
  title={Gradientless descent: High-dimensional zeroth-order optimization},
  author={Golovin, Daniel and Karro, John and Kochanski, Greg and Lee, Chansoo and Song, Xingyou and Zhang, Qiuyi},
  journal={arXiv preprint arXiv:1911.06317},
  year={2019}
}

@inproceedings{chen2017zoo,
  title={Zoo: Zeroth order optimization based black-box attacks to deep neural networks without training substitute models},
  author={Chen, Pin-Yu and Zhang, Huan and Sharma, Yash and Yi, Jinfeng and Hsieh, Cho-Jui},
  booktitle={Proceedings of the 10th ACM workshop on artificial intelligence and security},
  pages={15--26},
  year={2017}
}

@article{lesage2020second,
  title={Second-order online nonconvex optimization},
  author={Lesage-Landry, Antoine and Taylor, Joshua A and Shames, Iman},
  journal={IEEE Transactions on Automatic Control},
  volume={66},
  number={10},
  pages={4866--4872},
  year={2020},
  publisher={IEEE}
}

@article{lillicrap2015continuous,
  title={Continuous control with deep reinforcement learning},
  author={Lillicrap, TP},
  journal={arXiv preprint arXiv:1509.02971},
  year={2015}
}

@article{kraft1988software,
  title={A software package for sequential quadratic programming},
  author={Kraft, Dieter},
  journal={Forschungsbericht- Deutsche Forschungs- und Versuchsanstalt fur Luft- und Raumfahrt},
  year={1988}
}

@article{Powell1964AnEM,
  title={An efficient method for finding the minimum of a function of several variables without calculating derivatives},
  author={M. J. D. Powell},
  journal={Comput. J.},
  year={1964},
  volume={7},
  pages={155-162},
  url={https://api.semanticscholar.org/CorpusID:62756844}
}

@article{Nelder1965ASM,
  title={A Simplex Method for Function Minimization},
  author={John A. Nelder and Roger Mead},
  journal={Comput. J.},
  year={1965},
  volume={7},
  pages={308-313},
  url={https://api.semanticscholar.org/CorpusID:2208295}
}

@inproceedings{Nocedal2018NumericalO,
  title={Numerical Optimization},
  author={Jorge Nocedal and Stephen J. Wright},
  booktitle={Fundamental Statistical Inference},
  year={2018},
  url={https://api.semanticscholar.org/CorpusID:189864167}
}

@inproceedings{kunpeng2012descent,
  title={Descent search with mean direction evolution strategies based on GPU with CUDA},
  author={Kunpeng, Pang and Yugang, Li and Xiabi, Liu},
  booktitle={2012 13th International Conference on Parallel and Distributed Computing, Applications and Technologies},
  pages={298--304},
  year={2012},
  organization={IEEE}
}

@inproceedings{maheswaranathan2019guided,
  title={Guided evolutionary strategies: Augmenting random search with surrogate gradients},
  author={Maheswaranathan, Niru and Metz, Luke and Tucker, George and Choi, Dami and Sohl-Dickstein, Jascha},
  booktitle={International Conference on Machine Learning},
  pages={4264--4273},
  year={2019},
  organization={PMLR}
}

@inproceedings{liu2020self,
  title={Self-Guided Evolution Strategies with Historical Estimated Gradients.},
  author={Liu, Fei-Yu and Li, Zi-Niu and Qian, Chao},
  booktitle={IJCAI},
  pages={1474--1480},
  year={2020}
}

@article{braun2024stein,
  title={Stein Variational Evolution Strategies},
  author={Braun, Cornelius V and Lange, Robert T and Toussaint, Marc},
  journal={arXiv preprint arXiv:2410.10390},
  year={2024}
}

@article{liu2016stein,
  title={Stein variational gradient descent: A general purpose bayesian inference algorithm},
  author={Liu, Qiang and Wang, Dilin},
  journal={Advances in neural information processing systems},
  volume={29},
  year={2016}
}

\newpage
\appendix
\onecolumn
\section{Full EvoGrad\textsuperscript{2} Algorithm}

\begin{algorithm}
   \caption{EvoGrad\textsuperscript{2} (Evolutionary-Weighted Hessian-Gradient Learning)}
   \label{code:full OHGL}
\begin{algorithmic}
   \STATE \textbf{Input:} $x_0, \Omega, \alpha, \varepsilon_0, \gamma_{\alpha} < 1, \gamma_{\varepsilon} < 1, n_{\max}$
   \STATE $k \gets 0$
   \STATE $l \gets 0$
   \STATE $\Omega_j \gets \Omega$
   \STATE Map $W_0: \mathbb{R} \to \mathbb{R}$
   \STATE Initialize distribution \(\mathcal{N}\!\left(m_{CMA},\, \sigma^2C_{CMA}\right)\)

   \WHILE{$\text{Budget} > 0$}
      \STATE \textbf{Explore:}
      \STATE \hspace{1em} Generate samples $\mathcal{D}_k = \{ \tilde{x}_i \}_{i=1}^m, \tilde{x}_i \in V_{\varepsilon_l}(\tilde{x}_k)$
      \STATE \hspace{1em} Evaluate samples $y_i = f(\tilde{x}_i), \ i = 1, \dots, m$
      \STATE \hspace{1em} Add tuples to the replay buffer: $\mathcal{D} = \mathcal{D} \cup \mathcal{D}_k$
      
      \STATE \textbf{Create Dataset of Tuples:}
      \STATE \hspace{1em} $\mathcal{T} \gets \{ (\tilde{x}_i, \tilde{x}_j) \mid \|\tilde{x}_i - \tilde{x}_j\|_2 < \varepsilon, \forall i, j \}$
      \STATE \hspace{1em} Select random $m$ tuples from $\mathcal{T}$

      \STATE \textbf{Weighted Output Map:}
      \STATE \hspace{1em} Update \(\mathcal{N}\!\left(m_{CMA},\, \sigma^2C_{CMA}\right)\) with the new samples $\mathcal{D}_k$, using the CMA iteration update
      \STATE \hspace{1em} Assign weights $w_i = W_k(x_i) = P(x_i \mid \mathcal{D}, \mathcal{N}) = \frac{p_{\mathcal{N}}(x_i)}{\sum_{x’ \in \mathcal{D}} p_{\mathcal{N}}(x’)}$ to samples based on function values
      
      \STATE \textbf{Higher-Order Gradient and Hessian Learning:}
      \STATE \hspace{1em} Calculate the Hessian from the Jacobian of the network: $H_k(x) = \text{J}(g_{\theta_k}(x))$

      \STATE \hspace{1em} Optimize the network with GD using the formula in \ref{eq:taylor_loss}
      
      \STATE \textbf{Gradient Descent:}
      \STATE \hspace{1em} Update the current solution using second-order information:
      \STATE \hspace{1em} $x_{k+1} \gets x_k - \alpha \cdot \tilde{g}_{\theta_k}(x_k)$
      
      \IF{$f(\tilde{x}_{k+1}) > f(\tilde{x}_k)$ for $n_{\max}$ times in a row}
         \STATE Generate a new trust region: $\Omega_{l+1} = \gamma_{\alpha}|\Omega_l|$ with center at $x_{\text{best}}$
         \STATE Reinitialize distribution \(\mathcal{N}\!\left(m_{CMA},\, \sigma^2C_{CMA}\right)\)
         \STATE Map $W_l: \mathbb{R} \to \mathbb{R}$
         \STATE $l \gets l + 1$
         \STATE $\varepsilon_l \gets \gamma_{\varepsilon} \varepsilon_l$
      \ENDIF
      
      \IF{$f(\tilde{x}_k) < f(\tilde{x}_{\text{best}})$}
         \STATE $x_{\text{best}} = \tilde{x}_k$
      \ENDIF
      
      \STATE $k \gets k + 1; \text{Budget} \gets \text{Budget} - m$
   \ENDWHILE

   \STATE \textbf{Return:} $x_{\text{best}}$
\end{algorithmic}
\end{algorithm}
\textbf{Implementation notes}: We maintain a single reply buffer $\mathcal{D}$ containing all samples collected so far. At each iteration, we sample a new batch over a $\varepsilon$-radius ball. The reply buffer is used both to train the gradient network $g_{\theta_k}$ and the CMA's covariance matrix $C_{CMA}$, mean $m_{CMA}$, and std ${\sigma}$. 

The CMA update is extremely lightweight compared to training the neural network, since it consists only of closed-form mean and covariance updates. Thus, the computational overhead of maintaining the CMA distribution is negligible.

The CMA problem resets every trust region change (See in the algorithm, "Generate a new trust region" step), as we effectively solve a new sub-problem. Therefore, the distribution calculated up to this point is no longer relevant. To determine the weights, we compute the probability of each sample in $\mathcal{D}$ and measure the probability of each sample being selected under the distribution over $\mathcal{D}$.

In our implementation, we used the principles of Trust Region and Value normalization. For more details, see \cite{sarafian2020explicit} (Trust region, Value Normalizer, etc)

\section{Statistical Analysis}

\begin{table}[H]
    \centering
    \caption{p-value from t-test comparing EvoGrad\textsuperscript{2} and EvoGrad with each baseline algorithm.}
    \begin{tabular}{l|c|c}
        \textbf{Metric} & \textbf{EvoGrad} & \textbf{EvoGrad\textsuperscript{2}} \\
        \hline
        EvoGrad\textsuperscript{2}    & 1.0000 & 0.500000 \\
        EGL    & 0.0000001 & 0.0000001 \\
        IGL    & 0.0000001 & 0.0000001 \\
        CMA    & 0.0000001 & 0.0000001 \\
        CMA-L    & 0.0000001 & 0.0000001 \\
        CMA-T    & 0.0000001 & 0.0000001 \\
        BFGS    & 0.0000001 & 0.0000001 \\
        SLSQP    & 0.0000 & 0.0000 \\
        Nelder-Mean    & 0.0000 & 0.0000 \\
        Powell    & 0.0000 & 0.0000 \\
    \end{tabular}
    \label{tab:t_test_comparison}
\end{table}

\section{Generated code from Code Force Example}
\label{sec:code_force_generated_code}
\begin{figure}[t]
    \centering
    \includegraphics[width=\columnwidth]{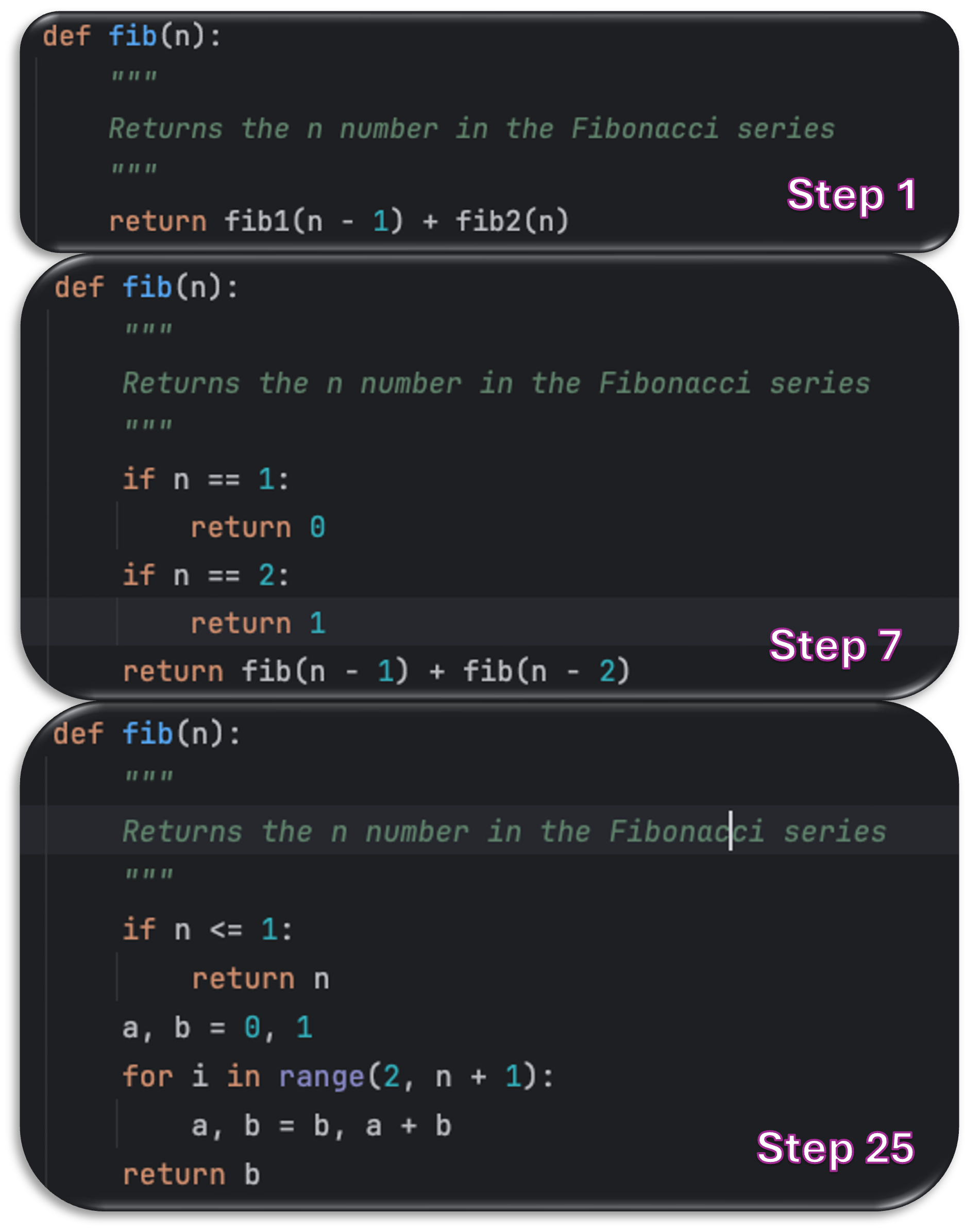}
    \caption{Generated code samples by EvoGrad\textsuperscript{2} algorithm, each step took 32 code samples for gradient computation}
    \label{fig:code_gen_sample1}
\end{figure}
\begin{figure}[H]
\centering
    \includegraphics[width=\textwidth]{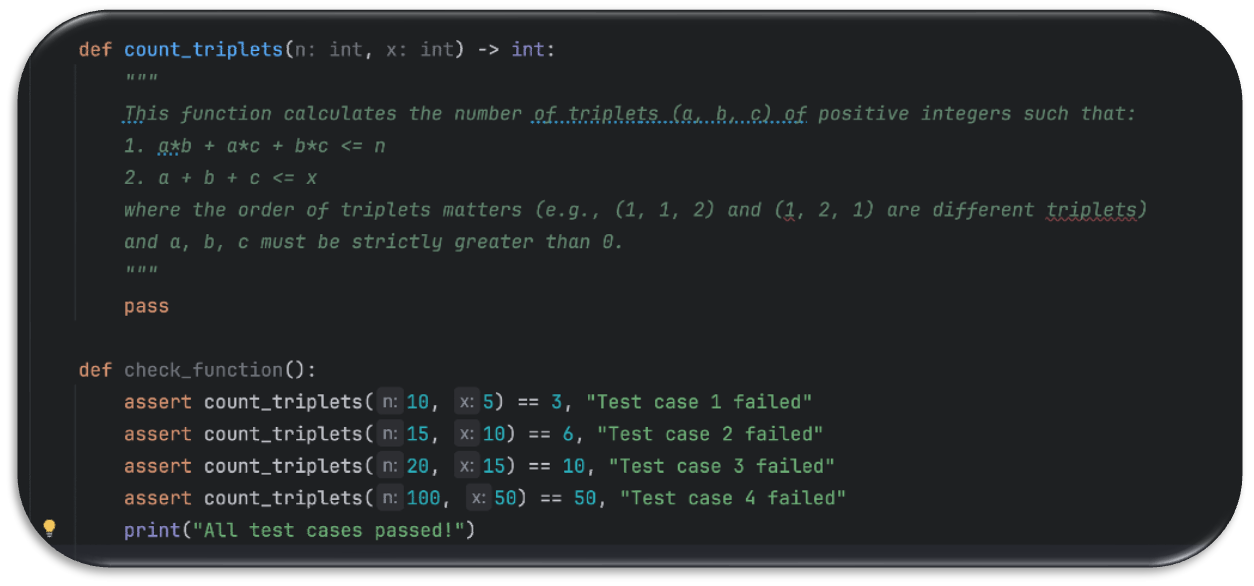}
    \captionof{figure}{Step 1}

\caption{Generated code samples by the algorithm}
\end{figure}

\begin{figure}[H]
\centering
        \includegraphics[width=\textwidth]{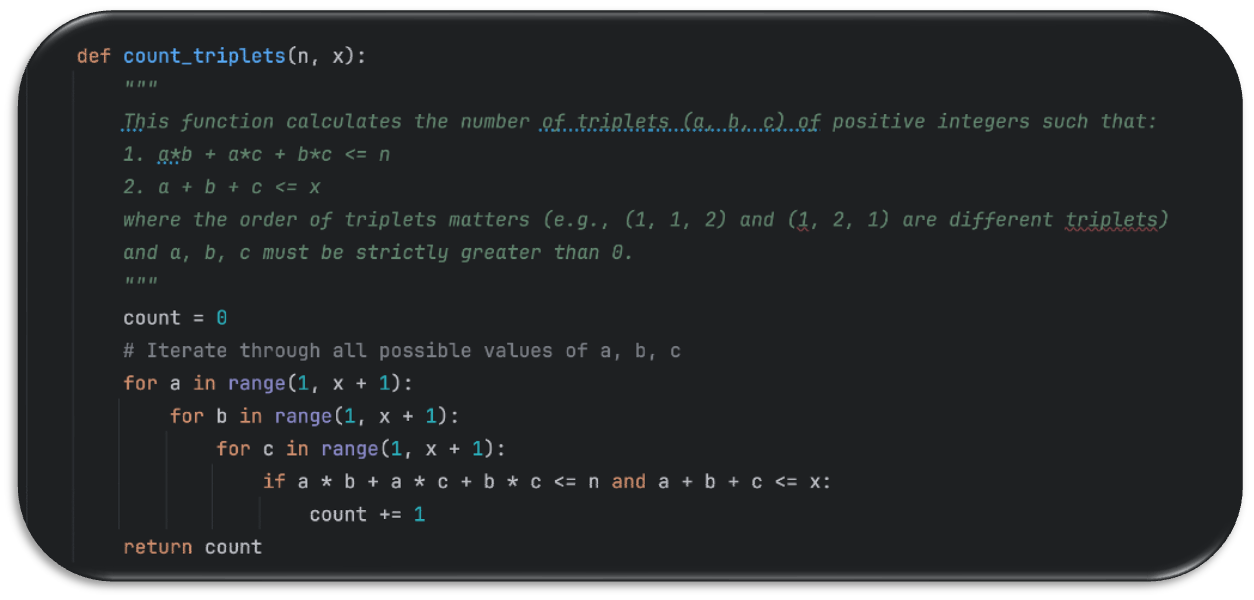}
    \captionof{figure}{Step 21}

\caption{Generated code samples by the algorithm}
\end{figure}

\begin{figure}[H]
\centering
        \includegraphics[width=\textwidth]{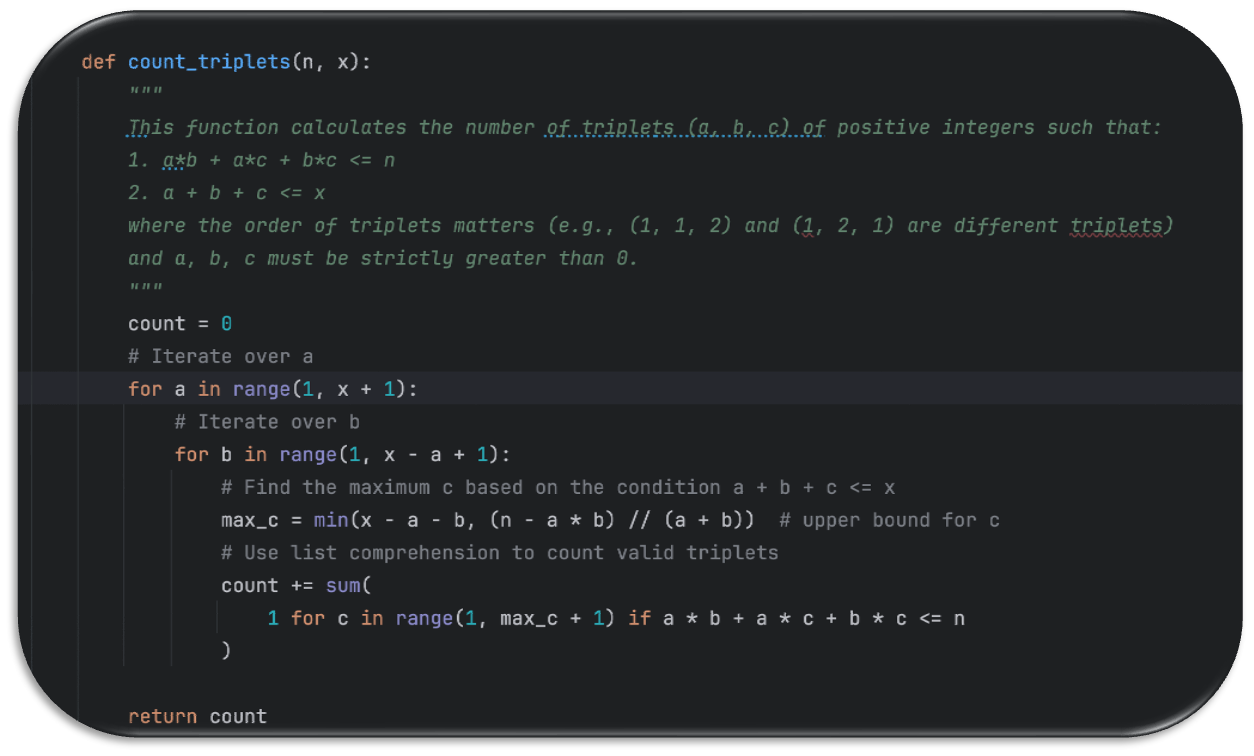}
    \captionof{figure}{Step 50}

\caption{Generated code samples by the algorithm}
\end{figure}

\section{Adaptive Sampling Size}
\label{sec:sampling_size_adapt}
Gradient learning algorithms collect samples to train the surrogate gradient model. While more samples can potentially lead to more accurate models, it is important to curtail the number of samples in each training iteration to be able to execute enough optimization steps before consuming the budget. On the other hand, one advantage that gradient learning has over direct objective learning is that we train the model with pairs of samples instead of single evaluation points, s.t. for a sample set of size $n_s$ we can draw as many as $n_p=n_s^2$ pairs to train our model. 

We empirically tested the optimal number of exploration sizes (i.e. $n_s$) in each training step and the optimal sample pair size ($n_p$) and found that a squared root profile is optimal for both hyperparameters. Therefore, we use the following equation to control these parameters:
\begin{equation}
    \begin{aligned}
        n_s & = 8 \cdot \left\lceil \sqrt{N} \right\rceil \\
        n_p & = 2000 \cdot \left\lceil \sqrt{N} \right\rceil \\
    \end{aligned}
\end{equation}
where $N$ is the problem size.

\section{Trust Region Management}
\label{sec:tr_management}
After finding the optimal solution inside the trust region, EGL shifts and scales down (i.e. shrinks) the trust region so that the optimal solution is centered at its origin. In case the decent path is long so that the minimum point is far away from the current trust region, this shrinking has the potential to slow down the learning rate and impede convergence.
To prevent unnecessary shrinking of the trust region, we distinguish between two convergence types: \textbf{interior convergence}: in this case, we shrink the trust region while moving its center and \textbf{boundary convergence} where we only shift the center without applying shrinking (When the algorithm reaches the edges of the TR). This adjustment prevents fast convergence of the step size to zero before being able to sufficiently explore the input domain. 
\newpage

\section{Benchmark Algorithms}
\label{sec:cma_benchmark_algorithm}

\begin{algorithm}
   \caption{CMA with a Trust Region}
   \label{alg:CMA_TR_ALG}
\begin{algorithmic}
   \STATE \textbf{Input:} $\texttt{total\_budget}, \texttt{budget} = 0, \texttt{start point } x, \delta = 1, \mu = 0, \gamma$
   \WHILE{$\texttt{budget} \leq \texttt{total\_budget}$}
      \STATE $\texttt{new\_generation}, \texttt{should\_stop} \gets \texttt{CMA}(x)$
      \STATE $\texttt{new\_generation} \gets \mu + \delta \cdot \texttt{new\_generation}$
      \STATE $\texttt{evaluations} \gets \texttt{space}(\texttt{new\_generation})$
      \STATE $\texttt{budget} \gets \texttt{budget} + \texttt{len}(\texttt{evaluations})$
      \STATE $\texttt{UpdateCovarMatrix}(\texttt{evaluations})$
      \IF{$\texttt{should\_stop}$}
         \STATE $\delta \gets \delta \cdot \gamma$
         \STATE $\mu \gets \arg\min_{x \in \texttt{new\_generation}}(\texttt{space}(x))$
      \ENDIF
   \ENDWHILE
\end{algorithmic}
\end{algorithm}

\section{Additional Empirical Results}
\begin{table}[H]
    \centering
    \caption{EGL Best Parameters} \label{tab:egl_best_params}
    \vspace{1ex}
    \begin{tabular}{lcp{5cm}}
        \toprule
        \textbf{Parameter} & \textbf{Value} & \textbf{Description} \\
        \midrule
        Exploration size & $8 \cdot \left\lceil \sqrt{\text{dimension}} \right\rceil$ & The number of iterations the algorithm will run. \\
        Maximum movement to shrink & $0.2 \cdot \sqrt{\text{dimension}}$ & How much distance the algorithm needs to cover to prevent shrinking when reaching convergence. \\
        Epsilon & $0.4 \cdot \sqrt{\text{dimension}}$ & Epsilon size to sample data. \\
        Epsilon Factor & 0.97 & How much we shrink the epsilon each iteration. \\
        Minimum epsilon & 0.0001 & The smallest epsilon we use. \\
        Database size & $20,000 \cdot \left\lceil \sqrt{\text{dimension}} \right\rceil$ & How many tuples we used to train for each iteration. \\
        Gradient network & dimension-10-15-10-dimension & What kind of network we use. \\
        Budget & 150,000 & How much budget the algorithm uses. \\
        \bottomrule
    \end{tabular}
    \vspace{1ex}
\end{table}

\section{CMA Versions}
\begin{figure}
    \centering
    \begin{tabular}{cc}
    \includegraphics[width=1.\textwidth]{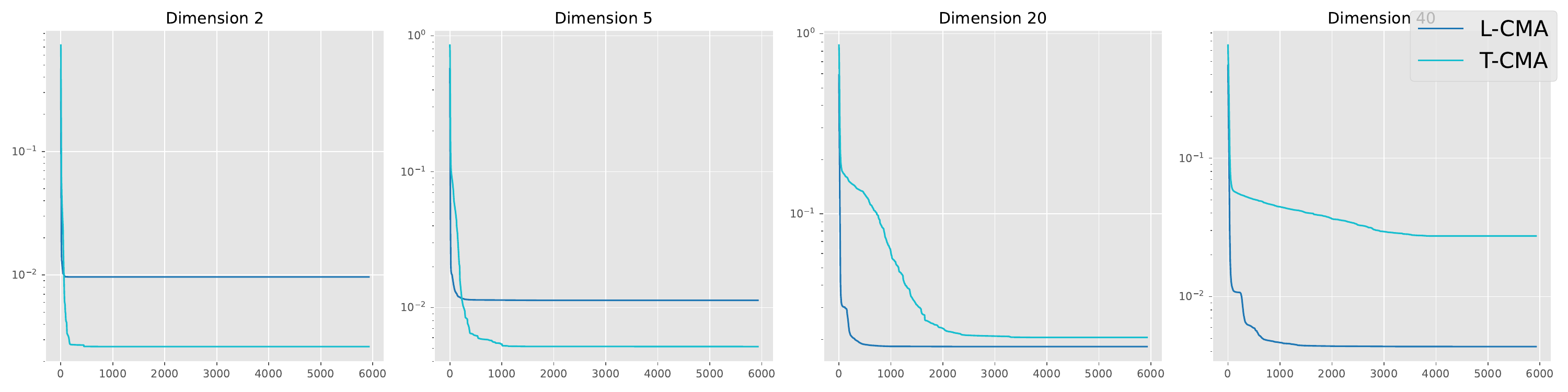}\hfill &
    \end{tabular}
    \caption{Comparing CMA with different trust regions types across different dimensions}
    \label{figure: cma_tr_dim_compare}
\end{figure}

The trust region is an extremely powerful tool. We map between the search space ($\Omega$) and trust region ($\Omega^*$) using the tanh function, which provides smooth transitions but exhibits logarithmic behavior at the edges, slowing large steps. While helpful for problems with many local minima, this behavior restricts CMA’s ability to take large steps, especially in high dimensions (see Figure \ref{figure: cma_tr_dim_compare}). We developed two CMA variants: one with linear TR (L-CMA) and one with tanh TR(T-CMA). The linear TR allowed larger steps, improving exploration, while the tanh TR limited performance.

\section{Adversarial Attack Implementation}
\label{sec:advarserial_atack_implementation}
Our adversarial attack leverages the flexibility of black-box optimization (BBO), which is not constrained by differentiability requirements on the objective function. This allows us to manipulate the search space freely. Specifically, our attack focuses on modifying the positions of $m$ fixed-size boxes within the image, along with the permutations applied to the pixels inside these boxes. This approach provides two key advantages: it enables us to limit the extent of the alterations we introduce to the image and simultaneously reduces the dimensionality of the search space $\Omega$.

The penalty function we use balances two competing goals: minimizing the cross-entropy loss to ensure misclassifications and limiting the perturbation magnitude using the mean squared error (MSE) loss. To achieve this balance, we define the following loss function:
\[
s(ce) = \sigma\left(b \cdot (\text{ce} - \epsilon)\right)
\]
\[
    Penalty_{mse}(\text{ce}, \text{mse}) \left(1 - s(ce)\right) \cdot \text{mse}^{n_1} + s(ce) \cdot \left( -\text{mse}^{n_2} \right)
\]
\[
    Penalty_{ce}(\text{ce}) \left(1 - s(ce))\right) \cdot e^{\text{ce} * n_1} + s(ce) \cdot \left( ln(\text{ce}^{n_2}) \right)
\]
\[
Penalty(x) = Penalty_{mse}(ce(x), mse(x)) - Penalty_{ce}(ce(x))
\]

In this formulation:
\begin{itemize}
    \item $\text{mse} \in [0, 1]$ and $\text{ce} \in (-\infty, \infty)$.
    \item The terms $n_1 \leq n_2$ control the trade-off between the cross-entropy and MSE loss slopes.
    \item $\epsilon$ defines the minimum required cross-entropy loss to maintain misclassification.
    \item The parameter $b$ governs the rate of transition between minimizing cross-entropy and MSE losses.
    \item The function $\sigma$ is the sigmoid function, which controls how much focus is given to minimizing cross-entropy loss over MSE loss as $\text{ce}$ increases.
\end{itemize}

This function balances the CR and the MSE, finding a minimum with CE to prevent detection by the classifier while minimizing the perturbation of the image.

\begin{figure}[ht]
\label{fig:egl_advarserial_attack_progression}
    \centering
    \includegraphics[width=\textwidth]{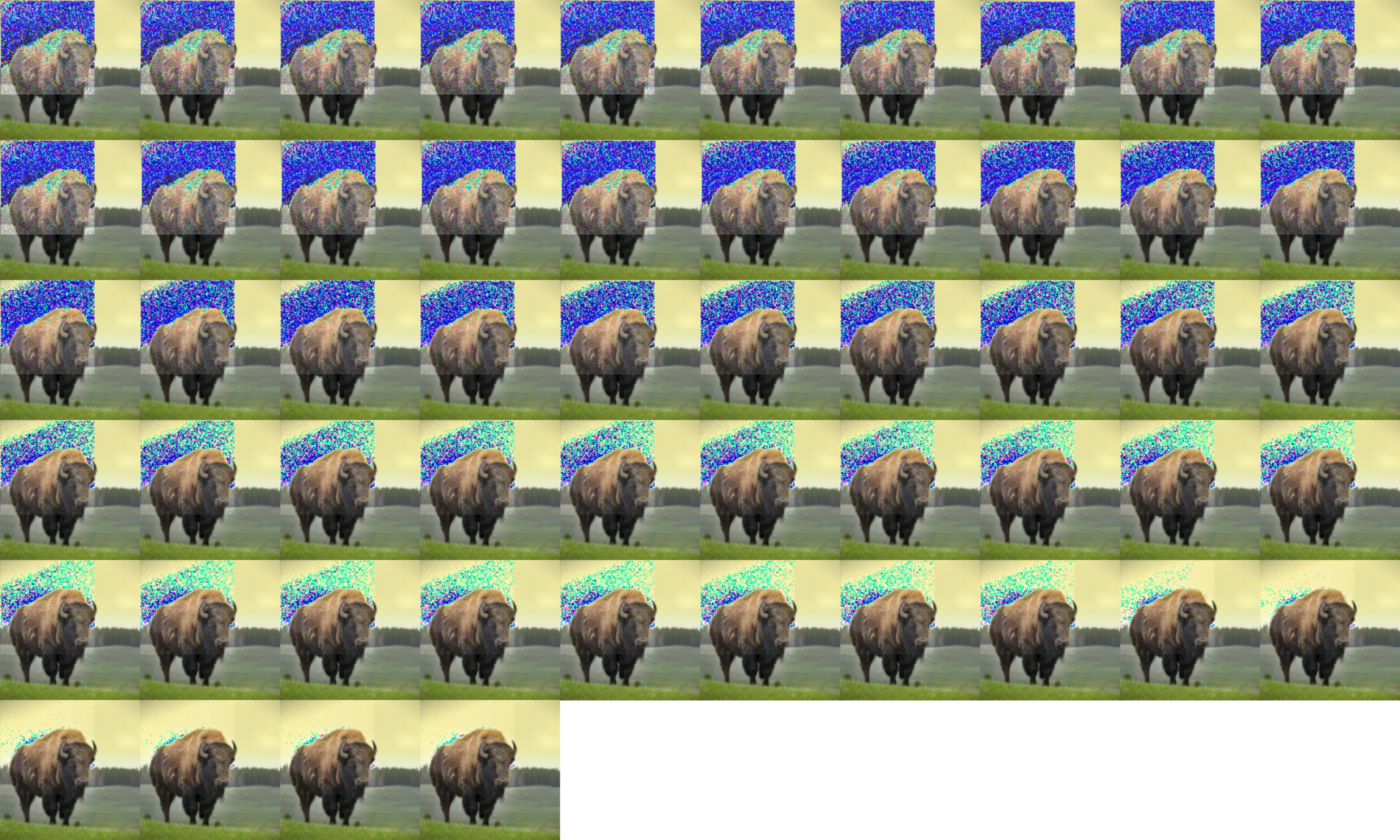}
    \caption{The search trajectory for an adversarial image generation with EvoGrad}
    \label{fig:model_images}
\end{figure}

\section{Full Robustness Experiment}

\begin{longtable}[H]{lcccc}
\caption{Comparison of Algorithm Versions on coco benchmark, comparing the error, the std, and the budget it takes to finish 99\% of the progress} \label{tab:algorithm_versions} \\
\toprule
\textbf{Version} & \textbf{Budget} & \textbf{Error} & \textbf{Std} & \textbf{Solved Problems} \\
\midrule
\endfirsthead
\toprule
\textbf{Version} & \textbf{Budget} & \textbf{Error} & \textbf{Std} & \textbf{Solved Problems} \\
\midrule
\endhead
\midrule
\multicolumn{5}{r}{{Continued on next page}} \\
\midrule
\endfoot
\bottomrule
\endlastfoot
\multicolumn{5}{c}{\textbf{networks}} \\
20 40 20 & 58447.91 & 0.020 & 0.031 & 0.82 \\
40 60 80 40 & 57960.44 & 0.0197 & 0.031 & 0.82 \\
60 80 60 & 58371.31 & 0.019 & 0.0311 & 0.82 \\
40 80 40 & 58159.34 & 0.021 & 0.031 & 0.82 \\
40 15 10 40 & 58667.00 & 0.021 & 0.031 & 0.82 \\
\midrule
\multicolumn{5}{c}{\textbf{Epsilon}} \\
0.1 & 39491.81 & 0.02 & 0.031 & 0.8 \\
0.4 & 48518.81 & 0.019 & 0.031 & 0.81 \\
0.5 & 49469.81 & 0.019 & 0.031 & 0.79 \\
0.6 & 54271.47 & 0.0203 & 0.031 & 0.78 \\
0.8 & 58667.00 & 0.0203 & 0.031 & 0.78 \\
\midrule
\multicolumn{5}{c}{\textbf{Epsilon Factor}} \\
0.8 & 39275.72 & 0.031 & 0.0365 & 0.74 \\
0.9 & 47930.81 & 0.02 & 0.0327 & 0.8 \\
0.95 & 53899.78 & 0.023 & 0.0326 & 0.78 \\
0.97 & 58667.00 & 0.027 & 0.0325 & 0.78 \\
0.99 & 63676.47 & 0.028 & 0.0326 & 0.77 \\
\midrule
\multicolumn{5}{c}{\textbf{Value Normalizer Outlier}} \\
0.01 & 57222.41 & 0.0059 & 0.0325 & 0.91 \\
0.05 & 57776.38 & 0.0060 & 0.0325 & 0.9 \\
0.1 & 58667.00 & 0.0057 & 0.0325 & 0.93 \\
0.2 & 58481.75 & 0.0061 & 0.0325 & 0.9 \\
0.3 & 58629.59 & 0.0063 & 0.0325 & 0.89 \\
\midrule
\multicolumn{5}{c}{\textbf{Trust Region Shrink Factor}} \\
0.99 & 103758.16 & 0.03 & 0.0433 & 0.69 \\
0.95 & 80660.69 & 0.024 & 0.0346 & 0.78 \\
0.9 & 58667.00 & 0.019 & 0.0325 & 0.8 \\
0.8 & 43503.22 & 0.022 & 0.0325 & 0.78 \\
0.7 & 39790.50 & 0.024 & 0.0326 & 0.76 \\
\midrule
\multicolumn{5}{c}{\textbf{Optimizer LR}} \\
0.001 & 101550.59 & 0.03 & 0.0364 & 0.68 \\
0.005 & 64988.66 & 0.025 & 0.0325 & 0.77 \\
0.01 & 58667.00 & 0.019 & 0.0325 & 0.8 \\
0.02 & 62405.84 & 0.018 & 0.0325 & 0.8 \\
0.03 & 72041.81 & 0.019 & 0.0325 & 0.8 \\
\midrule
\multicolumn{5}{c}{\textbf{Gradient LR}} \\
0.0001 & 63283.41 & 0.021 & 0.0325 & 0.78 \\
0.0005 & 59237.59 & 0.026 & 0.0420 & 0.76 \\
0.001 & 58667.00 & 0.019 & 0.0325 & 0.8 \\
0.002 & 58776.25 & 0.02 & 0.0325 & 0.8 \\
0.003 & 57934.91 & 0.021 & 0.0326 & 0.8 \\
\midrule
\bottomrule
\end{longtable}




\subsection{Theoretical Analysis}
\subsubsection{Mean Gradient Accuracy}
\label{sec:mean_gradient_acc_proof}
\begin{definition}
       The second-order mean gradient around x of radius $\epsilon > 0$:   \[
   \tilde{g}_{\varepsilon}(x) = \arg\min_{g \in \mathbb{R}^n} \int_{V_{\varepsilon}(x)} \left| g^\top \tau + \tfrac{1}{2} \tau^\top J(g) \tau - [ f(x + \tau) - f(x) ] \right|^2 d\tau.
   \]
\end{definition}
\begin{theorem} (Improved Controllable Accuracy):
\label{eq:controllable_acc_theorom}
For any twice differentiable function \( f \in \mathcal{C}^2 \), there exists \( \kappa_g(x) > 0 \) such that for any \( \varepsilon > 0 \), the second-order mean-gradient \( \tilde{g}_{\varepsilon}(x) \) satisfies
\[
\| \tilde{g}_{\varepsilon}(x) - \nabla f(x) \| \leq \kappa_g(x) \varepsilon^2 \quad \text{for all } x \in \Omega.
\]
\end{theorem}
\textit{Proof.}
   Since \( f \in \mathcal{C}^2 \), the Taylor expansion around \( x \) is\[
   f(x + \tau) = f(x) + \nabla f(x)^\top \tau + \tfrac{1}{2} \tau^\top H(x) \tau + R_{x}(\tau),
   \]
   where \( H(x) \) is the Hessian matrix at \( x \), and the remainder \( R_{x}(\tau) \) satisfies \[
   | R_{x}(\tau) | \leq \tfrac{1}{6} k_g \| \tau \|^3,
   \]
By the Definition of \(g_\epsilon\), an upper bound of \(\mathcal{L}(g_\epsilon(x))\)  is:   \[
   \mathcal{L}(g_\epsilon(x)) \leq \mathcal{L}(\nabla f(x)) = \int_{V_{\varepsilon}(x)} \Big| \nabla f(x)^\top \tau + \tfrac{1}{2} \tau^\top H(x) \tau - \big( f(x + \tau) - f(x) \big) \Big|^2 d\tau = \int_{V_{\varepsilon}(x)} | R_{x}(\tau) |^2 d\tau
\]
\[
   \leq \left( \tfrac{1}{6} k_g \right)^2 \int_{V_{\varepsilon}(x)} \| \tau \|^6 d\tau = \int_0^\varepsilon \int_{S_{n-1}} r^{n-1}r^6 drd\Omega= \left( \tfrac{1}{6} k_g \right)^2 \tfrac{1}{n+6}\omega_n \varepsilon^{n+6}.
\]
Where  \(\omega_n = \tfrac{2\pi^{n/2}}{\Gamma(n/2)}\) is the surface area of the unit sphere \(S^{n-1}\).
We now find a lower bound:
   
lets define for convenience \(\delta_g = (g_{\varepsilon}(x)-\nabla f(x))\) , \( \delta_H =  J(g_{\varepsilon}(x)) - H(x) \) 
\begin{align*}
    \mathcal{L}(g_{\varepsilon}(x)) &= \int_{V_{\varepsilon}(x)}|g_{\varepsilon}(x)\tau-\nabla f(x)\tau+\nabla f(x)\tau + \tfrac{1}{2}\tau  J(g_{\varepsilon}(x))\tau - \\& \tfrac{1}{2}\tau H(x) \tau + \tfrac{1}{2}\tau H(x)\tau - f(x+\tau)+f(x)|^2d\tau \\
& = \int_{V_{\varepsilon}(x)} (\delta_g^\top\tau + \tfrac{1}{2}\tau^\top\delta_H\tau - R_x(\tau))^2d\tau \\
& = \int_{V_{\varepsilon}(x)}((\delta_g^\top\tau)^2 + \tfrac{1}{4}(\tau^\top\delta_H\tau)^2 + R_x(\tau)^2 - 2\delta_g^\top\tau R_x(\tau) - \tau^\top\delta_H\tau  R_x(\tau) + \tau^\top\delta_g\tau^\top\delta_H\tau)d\tau\\
\end{align*}
Since \((a - b)^2 \geq 0\) we conclude that \( (\tfrac{1}{2}(\tau^\top\delta_H\tau) - R_x(\tau))^2 = \tfrac{1}{4}(\tau^\top\delta_H\tau)^2 + R_x(\tau)^2 - \tau^\top\delta_H\tau R_x(\tau) \geq 0\)
\begin{align*}
    \mathcal{L}(g_{\varepsilon}(x)) & = \int_{V_{\varepsilon}(x)}((\delta_g^\top\tau)^2 + \tfrac{1}{4}(\tau^\top\delta_H\tau)^2 + R_x(\tau)^2 - 2\delta_g^\top\tau R_x(\tau) - \tau^\top\delta_H\tau  R_x(\tau) + \tau^\top\delta_g\tau^\top\delta_H\tau)d\tau\\
    & \geq \int_{V_{\varepsilon}(x)}((\delta_g^\top\tau)^2 - 2\delta_g^\top\tau R_x(\tau) + \tau^\top\delta_g\tau^\top\delta_H\tau)d\tau\\
    & = \int_{V_{\varepsilon}(x)}(\delta_g^\top\tau)^2d\tau - 2\int_{V_{\varepsilon}(x)}\delta_g^\top\tau R_x(\tau)d\tau + \int_{V_{\varepsilon}(x)}\tau^\top\delta_g\tau^\top\delta_H\tau d\tau\\
    & = \int_{V_{\varepsilon}(x)}(\delta_g^\top\tau)^2d\tau - 2\int_{V_{\varepsilon}(x)}\delta_g^\top\tau R_x(\tau)d\tau + \int_{V_{\varepsilon}(x)}\tau^\top\delta_g\tau^\top\delta_H\tau d\tau\\
\end{align*}

We will split the equation into 3 components:
First \(A =  \int_{V_{\varepsilon}(x)}(\delta_g^\top\tau)^2d\tau\) ,  is we open this expression we see that
\[
(\delta_g^\top\tau)^2 = \sum_{i=1}^n \sum_{j=1}^n (\delta_g)_i (\delta_g)_j \tau_i \tau_j.
\]
Since \(V_{\varepsilon}(x)\) is symmetric around \(x\) odd moments of \(\tau\) integrate to zero, therefore
\begin{align*}
A &=  \int_{V_{\varepsilon}(x)}(\delta_g^\top\tau)^2d\tau = \int_{V_{\varepsilon}(x)}(\sum_{i=1}^n \sum_{j=1}^n (\delta_g)_i (\delta_g)_j \tau_i \tau_j) d\tau= \int_{V_{\varepsilon}(x)}(\sum_{i=1}^n  (\delta_g)_i^2 \tau_i^2 ) d\tau = \\
& = \sum_{i=1}^n  (\delta_g)_i^2\int_{V_{\varepsilon}(x)}\tau_i^2  d\tau =_* \frac{1}{n}\sum_{i=1}^n  (\delta_g)_i^2 \omega_n\int_{0}^\varepsilon r^{n+1}dr =  \sum_{i=1}^n  (\delta_g)_i^2 \omega_n \frac{1}{n(n+2)}\varepsilon^{n+2} =\\
& = \|\delta_g\|^2\omega_n \frac{1}{n(n+2)}\varepsilon^{n+2}
\end{align*}
Transition * stems from the fact that the integral over all\(\tau_i\) is the same due to the ball's symmetry. In other words 
\[\int_{V_{\varepsilon}(x)}\tau_i^2 d\tau = \frac{1}{n}\int_{V_{\varepsilon}(x)} n*\tau_i^2 d\tau  = \frac{1}{n}\int_{V_{\varepsilon}(x)} \sum_{j=0}^n\tau_j^2  d\tau = \frac{1}{n}\int_{V_{\varepsilon}(x)} \|\tau\| d\tau\]
and therefore
\[
\int_{V_{\varepsilon}(x)}\tau_i^2 d\tau = \frac{1}{n}\int_{V_{\varepsilon}(x)}\|\tau\|^2 d\tau = \frac{1}{n}\int_{0}^\varepsilon \int_{S_{n-1}}r^{n-1}r^2dr d\Omega = \frac{1}{n}\omega_n \int_{0}^\varepsilon r^{n+1}dr
\]

Let \(B = 2\int_{V_{\varepsilon}(x)}\delta_g^\top\tau R_x(\tau)d\tau\) 
\begin{align*}
    B &= 2\int_{V_{\varepsilon}(x)}\delta_g^\top\tau R_x(\tau)d\tau \leq 2\int_{V_{\varepsilon}(x)}\|\delta_g\|\cdot\|\tau\|\cdot\| R_x(\tau)\|d\tau  \leq 2\|\delta_g\|\int_{V_{\varepsilon}(x)}\|\tau\|\tfrac{1}{6}k_g\|\tau\|^3d\tau \\
    & \leq \tfrac{1}{3}k_g\|\delta_g\|\int_{V_{\varepsilon}(x)}\|\tau\|^4d\tau = \omega_n\int_0^\varepsilon r^{n-1}r^4dr = \tfrac{1}{3(n+4)}k_g\omega_n\|\delta_g\|\varepsilon^{n+4}
\end{align*}
Finally \(C = \int_{V_{\varepsilon}(x)}\tau^\top\delta_g\tau^\top\delta_H\tau d\tau\), we should remember the property derived from \(V_{\varepsilon}(x)\) symmetry. If we open the vector multiplication we get
\[
\tau^\top \delta_g \tau^\top \delta_H \tau = \sum_{i=1}^n \sum_{j=1}^n \sum_{k=1}^n \tau_i (\delta_g)_i (\delta_H)_{jk} \tau_j \tau_k = \sum_{i=1}^n \sum_{j=1}^n \sum_{k=1}^n (\delta_g)_i (\delta_H)_{jk} \tau_i \tau_j \tau_k.
\]
Here there is no way for an even moment of \(\tau\) to exist so \(C = 0\)
To sum this up, we get
\begin{align*}
    \mathcal{L}(g_{\varepsilon}(x)) &\geq A - B + C \geq \|\delta_g\|^2\omega_n \frac{1}{n(n+2)}\varepsilon^{n+2} - \tfrac{1}{3(n+4)}k_g\omega_n\|\delta_g\|\varepsilon^{n+4}
\end{align*}
We combine the lower and upper-bound 
\begin{align*}
    & \|\delta_g\|^2\omega_n \frac{1}{n(n+2)}\varepsilon^{n+2} - \tfrac{1}{3(n+4)}k_g\omega_n\|\delta_g\|\varepsilon^{n+4} \leq \left( \tfrac{1}{6} k_g \right)^2 \tfrac{1}{n+6}\omega_n \varepsilon^{n+6}\\
    & \|\delta_g\|^2 \frac{1}{n(n+2)} - \tfrac{1}{3(n+4)}k_g\|\delta_g\|\varepsilon^{2} - ( \tfrac{1}{6} k_g )^2 \tfrac{1}{n+6}\omega_n \varepsilon^{4} \leq 0\\
     &\|\delta_g\| \leq \frac{\tfrac{1}{3(n+4)}k_g\varepsilon^{2} + \sqrt{(\tfrac{1}{3(n+4)}k_g\varepsilon^{2})^2 + 4(\tfrac{1}{6} k_g )^2 \tfrac{1}{n+6}\varepsilon^{4} \frac{1}{n(n+2)}}}{2\frac{1}{n(n+2)}} = \frac{1}{3}\varepsilon^2 k_g\frac{\tfrac{1}{(n+4)} + \sqrt{\tfrac{1}{(n+4)^2} + \tfrac{1}{n+6} \frac{1}{n(n+2)}}}{2\frac{1}{n(n+2)}} \\
     & \|\delta_g\| \leq \frac{1}{3}\varepsilon^2 k_g \tfrac{\frac{1}{n+4} + \sqrt{\frac{4}{(n+4)^2}}}{2n(n+2)} = \frac{1}{2}n\varepsilon^2 k_g
\end{align*}

Now we show \(\|\delta_H\| \leq k_g\varepsilon\)
\begin{align*}
    \mathcal{L}(g_{\varepsilon}(x)) & = \int_{V_{\varepsilon}(x)}((\delta_g^\top\tau)^2 + \tfrac{1}{4}(\tau^\top\delta_H\tau)^2 + R_x(\tau)^2 - 2\delta_g^\top\tau R_x(\tau) - \tau^\top\delta_H\tau  R_x(\tau) + \tau^\top\delta_g\tau^\top\delta_H\tau)d\tau\\
    & \geq \int_{V_{\varepsilon}(x)}(\tfrac{1}{4}(\tau^\top\delta_H\tau)^2 - \tau^\top\delta_H\tau  R_x(\tau) + \tau^\top\delta_g\tau^\top\delta_H\tau)d\tau\\
    & = \int_{V_{\varepsilon}(x)}(\tfrac{1}{4}(\tau^\top\delta_H\tau)^2 - 
    \tau^\top\delta_H\tau  R_x(\tau))d\tau\\
    & = \tfrac{1}{4}\int_{V_{\varepsilon}(x)}(\tau^\top\delta_H\tau)^2d\tau - \int_{V_{\varepsilon}(x)}\tau^\top\delta_H\tau  R_x(\tau)d\tau\\
\end{align*}
\begin{align*}
    A &= \int_{V_{\varepsilon}(x)}(\tau^\top\delta_H\tau)^2 d\tau=  \int_{V_{\varepsilon}(x)} \sum_{i,j,k,l=0}^n \tau_i \tau_j \tau_k \tau_l (\delta_H)_{ij} (\delta_H)_{kl} d\tau \\
    & =_* \int_{V_{\varepsilon}(x)} \sum_{i=0}^n \tau_i^4 (\delta_H)_{ii}^2 d\tau + 2\int_{V_{\varepsilon}(x)} \sum_{i \neq j}^n \tau_i^2 \tau_j^2 (\delta_H)_{ij}^2  d\tau + \int_{V_{\varepsilon}(x)} \sum_{i\neq j}^n \tau_i^2 \tau_j^2 (\delta_H)_{ii} (\delta_H)_{jj} d\tau \\
    & = \sum_{i=0}^n (\delta_H)_{ii}^2 \omega_n \frac{\varepsilon^{n+4}}{n+4}\frac{3}{n(n+2)} + 2\sum_{i \neq j}^n (\delta_H)_{ij}^2  \omega_n \frac{\varepsilon^{n+4}}{n+4}\frac{1}{n(n+2)} + \sum_{i\neq j}^n (\delta_H)_{ii} (\delta_H)_{jj}  \omega_n \frac{\varepsilon^{n+4}}{n+4}\frac{1}{n(n+2)} \\
\end{align*}
Where * is due to the symmetry of the ball, any odd moments of \(\tau\) are equal to 0

Let's denote
\[
M =  \omega_n \frac{\varepsilon^{n+4}}{n+4}\frac{1}{n(n+2)}
\]
\[
S_{diag} = \sum_{i=0}^n (\delta_H)_{ii}^2
\]
\[
S_{off-diag} = \sum_{i\neq j} (\delta_H)_{ij}^2
\]
\[
T  = \sum_{i = 0}^n (\delta_H)_{ii}
\]
We notice that
\[
\sum_{i\neq j}^n (\delta_H)_{ii} (\delta_H)_{jj}  =T^2 - S_{diag}
\]
\[
\|\delta_H\|^2 = S_{diag} + S_{off-diag}
\]
\begin{align*}
    A &= 3S_{diag} \cdot M + 2 M\cdot S_{off-diag} + (T^2 - S_{diag})M = 2 (S_{diag} + S_{off-diag}) + M \cdot T^2 \geq M \|\delta_H\|^2
\end{align*}

\begin{align*}
    B = \int_{V_{\varepsilon}(x)}\tau^\top\delta_H\tau  R_x(\tau)d\tau \leq \frac{1}{6}k_g\|\delta_H\|\int_{V_{\varepsilon}(x)}\|\tau\|^5d\tau = \|\delta_H\|\tfrac{1}{6(n+5)}k_g\omega_n\varepsilon^{n+5}
\end{align*}

\begin{align*}
  &\left( \tfrac{1}{6} k_g \right)^2 \tfrac{1}{n+6}\omega_n \varepsilon^{n+6}  \geq \mathcal{L}(g_{\varepsilon}(x)) \geq \frac{1}{4}A - B \geq  \omega_n \frac{\varepsilon^{n+4}}{n+4}\frac{1}{n(n+2)} \|\delta_H\|^2 - \tfrac{1}{6(n+5)}k_g\omega_n\varepsilon^{n+5}\|\delta_H\|\\
 & \frac{1}{n+4}\frac{1}{n(n+2)} \|\delta_H\|^2 - \tfrac{1}{6(n+5)}k_g\varepsilon\|\delta_H\| - \left( \tfrac{1}{6} k_g \right)^2 \tfrac{1}{n+6} \varepsilon^{2} \leq 0 \\
 & \|\delta_H\| \leq \frac{\frac{1}{6(n+5)} k_g \varepsilon \pm \sqrt{\frac{k_g^2 \varepsilon^2}{36} \left(\frac{1}{(n+5)^2} + \frac{4}{(n+4)n(n+2)(n+6)}\right)}}{\frac{2}{n+4} \frac{1}{n(n+2)}} \leq \tfrac{1}{6}k_g \varepsilon n^2
\end{align*}

\subsubsection{Evolutionary Gradient Accuracy}
\label{sec:optimistic_error_proof}
\begin{definition}
       The evolutionary gradient around x of radius $\epsilon > 0$:   \[
   \tilde{g}_{\varepsilon}(x) = \arg\min_{g \in \mathbb{R}^n} \int_{V_{\varepsilon}(x)} w(\tau)\left| g^\top \tau  - [ f(x + \tau) - f(x) ] \right|^2 d\tau.
   \]
\end{definition}
\begin{theorem} (Evolutionary gradient controllable Accuracy):
\label{eq:ogl_controllable_acc_proof}
For any twice differentiable function \( f \in \mathcal{C}^2 \), there exists \( \kappa_g(x) > 0 \) such that for any \( \varepsilon > 0 \), the second-order mean-gradient \( \tilde{g}_{\varepsilon}(x) \) satisfies
\[
\| \tilde{g}_{\varepsilon}(x) - \nabla f(x) \| \leq \kappa_g(x) \varepsilon \quad \text{for all } x \in \Omega.
\]
\end{theorem}
\textit{Proof.}
   Since \( f \in \mathcal{C}^2 \)the Taylor expansion around \( x \) is\[
   f(x + \tau) = f(x) + \nabla f(x)^\top \tau + R_{x}(\tau),
   \]
   where \( H(x) \) is the Hessian matrix at \( x \), and the remainder \( R_{x}(\tau) \) satisfies \[
   | R_{x}(\tau) | \leq \tfrac{1}{2} k_g \| \tau \|^2,
   \]
By definition \(g_\epsilon\), an upper bound  \(\mathcal{L}(g_\epsilon(x))\)  is:   \[
   \mathcal{L}(g_\epsilon(x)) \leq \mathcal{L}(\nabla f(x)) = \int_{V_{\varepsilon}(x)} \Big| \nabla f(x)^\top \tau - \big( f(x + \tau) - f(x) \big) \Big|^2 d\tau = \int_{V_{\varepsilon}(x)} | R_{x}(\tau) |^2 d\tau
\]
\[
   \leq \left( \tfrac{1}{2} k_g \right)^2 \int_{V_{\varepsilon}(x)} \| \tau \|^4 d\tau = \left( \tfrac{1}{2} k_g \right)^2|V_1(x)| \varepsilon^{n+4}.
\]

\textit{Proof.}
The evolutionary mean gradient around x of radius $\epsilon > 0$:   \[
   \tilde{g}_{\varepsilon}(x) = \arg\min_{g \in \mathbb{R}^n} \int_{V_{\varepsilon}(x)} w(\tau)\left| g^\top \tau  - [ f(x + \tau) - f(x) ] \right|^2 d\tau.
   \]
\begin{align*}
       \mathcal{L}(g_{\varepsilon}(x)) & = \int_{V_{\varepsilon}(x)} w(\tau) (\delta_g^\top\tau - R_x(\tau))^2 d\tau \geq \int_{V_{\varepsilon}(x)} w(\tau) (\delta_g^\top\tau)^2 - 2\cdot w(\tau)\delta_g^\top R_x(\tau) + w(\tau) R_x(\tau)^2 d\tau \\
       & = \int_{V_{\varepsilon}(x)} w(\tau) (\delta_g^\top\tau)^2 - \int_{V_{\varepsilon}(x)} 2\cdot w(\tau)\delta_g^\top\cdot R_x(\tau) d\tau + \int_{V_{\varepsilon}(x)} w(\tau) R_x(\tau)^2 d\tau \\
       & \geq \int_{V_{\varepsilon}(x)} w(\tau) (\delta_g^\top\tau)^2 - 2\int_{V_{\varepsilon}(x)} w(\tau)\delta_g^\top\cdot R_x(\tau) d\tau \\
   \end{align*}
   Since \(V_{\varepsilon}(x)\) is symmetric around \(x\) odd moments of \(\tau\) integrate to zero, therefore
\begin{align*}
A &=\; \int_{V_{\varepsilon}(x)} w(\tau)\,\big(\delta_g^\top \tau\big)^2d\tau 
\;\ge\; W_{\min}\int_{V_{\varepsilon}(x)} \big(\delta_g^\top \tau\big)^2d\tau =\; W_{\min}\,\delta_g^\top\!\Big(\int_{V_{\varepsilon}(x)} \tau\,\tau^\top d\tau\Big)\delta_g\ \\& = \; W_{\min}\delta_g^\top\!\Big(\int_{V_{\varepsilon}(x)} \tau\,\tau^\top d\tau\Big)\delta_g \;=W_{\min}\; C_n\,\varepsilon^{\,n+2}\,\|\delta_g\|^2 \;=\; \frac{W_{\min}}{\,n+2\,}\,|V_1(x)|\,\varepsilon^{\,n+2}\,\|\delta_g\|^2
\end{align*}
Let \(B = 2\int_{V_{\varepsilon}(x)}w(\tau)\delta_g^\top\tau R_x(\tau)d\tau\) 
\begin{align*}
    B &= 2\int_{V_{\varepsilon}(x)}w(\tau)\delta_g^\top\tau R_x(\tau)d\tau \leq 2\int_{V_{\varepsilon}(x)}w(\tau)\|\delta_g\|\cdot\|\tau\|\cdot R_x(\tau)d\tau  \leq 2\|\delta_g\|\int_{V_{\varepsilon}(x)}w(\tau)\|\tau||\tfrac{1}{2}k_g\tau^2d\tau \\
    & \leq \tfrac{1}{3}k_g\|\delta_g\|\int_{V_{\varepsilon}(x)}w(\tau)\|\tau\|^3d\tau = k_g||\delta_g||\varepsilon^{n+3}|V_1(x)|W_u
\end{align*}
To sum this up, we get
\begin{align*}
    \mathcal{L}(g_{\varepsilon}(x)) &\geq A - B  \geq \frac{W_{\min}}{\,n+2\,}\,|V_1(x)|\,\varepsilon^{\,n+2}\,\|\delta_g\|^2 - k_g||\delta_g||\varepsilon^{n+3}|V_1(x)|W_u
\end{align*}
We combine the lower and upper-bound 
\begin{align*}
    & \frac{W_{\min}}{\,n+2\,}\,|V_1(x)|\,\varepsilon^{\,n+2}\,\|\delta_g\|^2 - k_g||\delta_g||\varepsilon^{n+3}|V_1(x)|W_u \leq ( \tfrac{1}{2} k_g )^2|V_1(x)| \varepsilon^{n+4}\\ \\
    & \|\delta_g\|^2 \frac{W_{\min}}{\,n+2\,}\, - k_g\|\delta_g\|\varepsilon W_u  - ( \tfrac{1}{2} k_g )^2 \varepsilon^{2}\leq 0\\
    & \|\delta_g\| \leq \frac{k_g \varepsilon W_u \pm \sqrt{k_g^2 \varepsilon^2 W_u^2 + k_g^2 \varepsilon^2 \cdot \frac{W_{\min}}{n+2}}}{2 \cdot \frac{W_{\min}}{n+2} } = \varepsilon k_g \cdot \frac{W_u \pm \sqrt{W_u^2 + \cdot \frac{W_{\min}}{n+2}}}{2\cdot \frac{W_{\min}}{n+2}}
\end{align*}

\subsubsection{Convergence Analysis}
\label{sec:convergence_proof}
\begin{corollary}
\label{eq:convergence_base_collary}
Let $f:\Omega\to\mathbb{R}$ be a convex function with Lipschitz continuous gradient, i.e. $f\in\mathcal{C}^{+1}$ and a Lipschitz constant $\kappa_f$ and let $f(x^{*})$ be its optimal value. Suppose a controllable mean-gradient model $g_{\varepsilon}$ with error constant $\kappa_g$, i.e. $\|g_{\varepsilon}(x) - \nabla f(x)\| \leq t(\varepsilon) \kappa_g $, the gradient descent iteration $x_{k+1} = x_{k} - \alpha g_{\varepsilon}(x_k)$ with a sufficiently small $\alpha$ s.t. $\alpha \leq \min(\frac{1}{\kappa_g}, \frac{1}{\kappa_f})$ guarantees:
\begin{enumerate}
    \item For $t(\varepsilon) \leq \frac{\|\nabla f(x)\|}{5\alpha}$, monotonically decreasing steps s.t. $f(x_{k+1}) \leq f(x_k) -  2.25 \frac{t(\varepsilon)}{\alpha}$.
    \item After a finite number of iterations, the descent process yields $x^{\star}$ s.t. $\|\nabla f(x^{\star})\| \leq \frac{5t(\varepsilon)}{\alpha}$.
\end{enumerate}
\end{corollary}
\textit{Proof.}
For a convex function with Lipschitz continuous gradient, the following inequality holds for all $x_k$
\begin{equation}
    f(x) \leq f(x_k) + (x - x_k) \cdot \nabla f(x_k) + \frac{1}{2}\kappa_f\|x - x_k\|^2
\end{equation}
Plugging in the iteration update $x_{k+1} = x_k - \alpha g_{\varepsilon}(x_k) $ we get
\begin{equation}
    f(x_{k+1}) \leq f(x_k) - \alpha g_{\varepsilon}(x_k) \cdot \nabla f(x_k) + \alpha^2 \frac{1}{2}\kappa_f\|g_{\varepsilon}(x_k)\|^2
\end{equation}
For a controllable mean-gradient we can write $\|g_{\varepsilon}(x) - \nabla f(x)\| \leq t(\varepsilon) \kappa_g $, therefore we can write $g_{\varepsilon}(x) = \nabla f(x) + t(\varepsilon) \kappa_g \xi(x)$ s.t. $\|\xi(x)\| \leq 1$ so the inequality is
\begin{equation}
    f(x_{k+1}) \leq f(x_k) - \alpha \|\nabla f(x_k)\|^2 - \alpha t(\varepsilon) \kappa_g \xi(x_k) \cdot \nabla f(x_k) + \alpha^2 \frac{1}{2}\kappa_f\|\nabla f(x) + t(\varepsilon) \kappa_g \xi(x)\|^2
\end{equation}
Using the equality $\|a+b\|^2 = \|a\|^2 + 2 a \cdot b + \|b\|^2$ and the Cauchy-Schwartz inequality inequality $a \cdot b\leq\|a\|\|b\|$ we can write

\begin{equation} 
\begin{aligned}
    f(x_{k+1}) & \leq f(x_k) - \alpha |\nabla f(x_k)|^2 - \alpha t(\varepsilon) \kappa_g |\xi(x_k)| \cdot |\nabla f(x_k)| \\ & \quad + \frac{\alpha^2}{2} \kappa_f \big( |\nabla f(x_k)|^2 + 2 t(\varepsilon) \kappa_g |\xi(x_k)| \cdot |\nabla f(x_k)| + t(\varepsilon)^2 \kappa_g^2 |\xi(x_k)|^2 \big) \\ &\leq f(x_k) - \alpha |\nabla f(x_k)|^2 - \alpha t(\varepsilon) \kappa_g |\nabla f(x_k)| \\ &\quad + \frac{\alpha^2}{2} \kappa_f |\nabla f(x_k)|^2 + \alpha^2 \kappa_f t(\varepsilon) \kappa_g |\nabla f(x_k)| + \frac{\alpha^2}{2} t(\varepsilon)^2 \kappa_f \kappa_g^2 \\ &= f(x_k) - \left( \alpha - \frac{\kappa_f}{2} \alpha^2 \right) |\nabla f(x_k)|^2 + \left( -\alpha + \kappa_f \alpha^2 \right) t(\varepsilon) \kappa_g |\nabla f(x_k)| \\ &\quad + \frac{\kappa_f}{2} \alpha^2 t(\varepsilon)^2 \kappa_g^2 
\end{aligned}
\end{equation}
Using the requirement  $\alpha \leq \min(\frac{1}{\kappa_g}, \frac{1}{\kappa_f})$ it follows that $\alpha \kappa_g \leq 1$ and $\alpha \kappa_f \leq 1$ so
\begin{equation}
\begin{aligned}
    f(x_{k+1}) & \leq f(x_k) - \left( \alpha - \frac{\kappa_f}{2} \alpha^2 \right) |\nabla f(x_k)|^2 + \left( -\alpha + \kappa_f \alpha^2 \right) t(\varepsilon) \kappa_g |\nabla f(x_k)| \\ &\quad + \frac{\kappa_f}{2} \alpha^2 t(\varepsilon)^2 \kappa_g^2 \\ &\leq f(x_k) - \frac{\alpha}{2} |\nabla f(x_k)|^2 + \frac{\kappa_f}{2} \alpha^2 t(\varepsilon)^2 \kappa_g^2 \\ & \leq f(x_k) - \frac{\alpha}{2} |\nabla f(x_k)|^2 + \frac{\kappa_g }{2} t(\varepsilon)^2 
\end{aligned} 
\end{equation}
Now, for $x$ s.t. $\|\nabla f(x)\| \geq \frac{5t(\varepsilon)}{\alpha}$ then $t(\varepsilon) \leq \|\nabla f(x)\|\tfrac{\alpha}{5}$. Plugging it into our inequality, we obtain
\begin{equation}
\begin{aligned}
    f(x_{k+1}) & \leq f(x_k) - \frac{\alpha}{2} \|\nabla f(x_k)\|^2 + \frac{\alpha^2}{50} \kappa_g \|\nabla f(x_k)\|^2 \\ & \leq f(x_k) - \frac{\alpha}{2} \|\nabla f(x_k)\|^2 + \frac{\alpha}{50} \|\nabla f(x_k)\|^2 \\ &= f(x_k) - 0.48 \alpha \|\nabla f(x_k)\|^2 \\ &\leq f(x_k) - \frac{12 t(\varepsilon)^2}{\alpha}
\end{aligned}
\end{equation}

Therefore, for all $x$ s.t., $\|\nabla f(x)\| \geq \frac{5\varepsilon^2}{\alpha}$ we have a monotonically decreasing step with finite size improvement, after a finite number of steps we obtain $x^{\star}$ for which $\|\nabla f(x^{\star})\| \leq \frac{5t(\varepsilon)}{\alpha}$.

\begin{corollary}
\label{eq:hgl_convergence_proof}
Let $f:\Omega\to\mathbb{R}$ be a convex function with Lipschitz continuous gradient, i.e. $f\in\mathcal{C}^{+1}$ and a Lipschitz constant $\kappa_f$ and let $f(x^{*})$ be its optimal value. Suppose a controllable mean-gradient model $g_{\varepsilon}$ with error constant $\kappa_g$, i.e. $\|g_{\varepsilon}(x) - \nabla f(x)\| \leq \varepsilon^2 \kappa_g $, the gradient descent iteration $x_{k+1} = x_{k} - \alpha g_{\varepsilon}(x_k)$ with a sufficiently small $\alpha$ s.t. $\alpha \leq \min(\frac{1}{\kappa_g}, \frac{1}{\kappa_f})$ guarantees:
\begin{enumerate}
    \item For $\varepsilon^2 \leq \frac{\|\nabla f(x)\|}{5\alpha}$, monotonically decreasing steps s.t. $f(x_{k+1}) \leq f(x_k) -  2.25 \frac{\varepsilon^2}{\alpha}$.
    \item After a finite number of iterations, the descent process yields $x^{\star}$ s.t. $\|\nabla f(x^{\star})\| \leq \frac{5\varepsilon^2}{\alpha}$.
\end{enumerate}

\end{corollary}
\textit{Proof.}
This is derived naturally from corollary \ref{eq:convergence_base_collary} and theorem \ref{eq:controllable_acc_theorom}

\begin{corollary}
\label{eq:ogl_convergence_proof}
Let $f:\Omega\to\mathbb{R}$ be a convex function with Lipschitz continuous gradient, i.e. $f\in\mathcal{C}^{+1}$ and a Lipschitz constant $\kappa_f$ and let $f(x^{*})$ be its optimal value. Suppose a controllable mean-gradient model $g_{\varepsilon}$ with error constant $\kappa_g$, i.e. $\|g_{\varepsilon}(x) - \nabla f(x)\| \leq \varepsilon \kappa_g $, the gradient descent iteration $x_{k+1} = x_{k} - \alpha g_{\varepsilon}(x_k)$ with a sufficiently small $\alpha$ s.t. $\alpha \leq \min(\frac{1}{\kappa_g}, \frac{1}{\kappa_f})$ guarantees:
\begin{enumerate}
    \item For $\varepsilon \leq \frac{\|\nabla f(x)\|}{5\alpha}$, monotonically decreasing steps s.t. $f(x_{k+1}) \leq f(x_k) -  2.25 \frac{\varepsilon}{\alpha}$.
    \item After a finite number of iterations, the descent process yields $x^{\star}$ s.t. $\|\nabla f(x^{\star})\| \leq \frac{5\varepsilon}{\alpha}$.
\end{enumerate}

\end{corollary}
\textit{Proof.}
This is derived naturally from corollary \ref{eq:convergence_base_collary} and theorem \ref{eq:ogl_controllable_acc_proof}

\end{document}